\documentclass[10pt, twocolumn, letterpaper]{article}

\usepackage[pagenumbers]{cvpr} 

\usepackage{graphicx}
\usepackage{amsmath}
\usepackage{amssymb}
\usepackage{booktabs}
\usepackage{hyperref}
\usepackage{graphicx} 

\ifdefined\siggraph
\usepackage{times}
\fi

\usepackage{color}
\usepackage{ifthen}
\usepackage{float}
\usepackage{alltt}
\usepackage{newlfont} 
\usepackage{array}
\usepackage{wrapfig}
\usepackage{booktabs}
\usepackage{multirow}
\usepackage{amsfonts}
\usepackage{dsfont}
\usepackage{xcolor}
\usepackage[linesnumbered,ruled,vlined]{algorithm2e}

\usepackage{mleftright}

\usepackage{soul}
 
\usepackage{epsfig}
\usepackage{graphicx}
\usepackage{amsmath}
\usepackage{amssymb}
\DeclareMathOperator*{\argmax}{arg\,max}

\newcommand{\nothing}[1]{}

\definecolor{AudioColor}{rgb}{0.56,0.34,0.62}

\definecolor{DeadlineColor}{rgb}{0.9,0.4,0} 

\definecolor{figred}{rgb}{1,0,0}
\definecolor{figgreen}{rgb}{0,0.6,0}
\definecolor{figblue}{rgb}{0,0,1}
\definecolor{figpink}{rgb}{1,0.63,0.63}

\newcolumntype{C}[1]{>{\centering}m{#1}}

\newcounter{pccount}
\floatstyle{plain}

\newcommand{\filename}[1]{\url{#1}}
\newcommand{\foldername}[1]{\url{#1}}

\newcommand{\shortname}{DiaNA}

\usepackage{hyperref}
\hypersetup{
	colorlinks=true,
	linkcolor=red,
	filecolor=blue,      
	urlcolor=red,
	citecolor=green,
}

\usepackage[capitalize]{cleveref}
\crefname{section}{Sec.}{Secs.}
\Crefname{section}{Section}{Sections}
\Crefname{table}{Table}{Tables}
\crefname{table}{Tab.}{Tabs.}

\usepackage[linesnumbered,ruled,vlined]{algorithm2e}

\SetKwInput{KwInput}{Input}                
\SetKwInput{KwOutput}{Output}              

\def\cvprPaperID{000} 
\def\confName{CVPR}
\def\confYear{2023}

\begin{document}
\pagestyle{plain}
\title{Divide and Adapt: Active Domain Adaptation via Customized Learning}

\author{Duojun Huang\textsuperscript{1,2}\quad Jichang Li\textsuperscript{3}\quad Weikai Chen\textsuperscript{4}\quad Junshi Huang\textsuperscript{5}\quad Zhenhua Chai\textsuperscript{5}\quad Guanbin Li\textsuperscript{1,2}\footnotemark[2]\\
\textsuperscript{1}School of Computer Science and Engineering, Sun Yat-sen University, Guangzhou, China \\
\textsuperscript{2}Research Institute, Sun Yat-sen University, Shenzhen, China \\
\textsuperscript{3}The University of Hong Kong \quad \textsuperscript{4}Tencent America \quad \textsuperscript{5}Meituan\\
{\tt\small huangdj9@mail2.sysu.edu.cn, csjcli@connect.hku.hk, chenwk891@gmail.com}\\
{\tt\small \{huangjunshi,chaizhenhua\}@meituan.com, liguanbin@mail.sysu.edu.cn}
}

\maketitle

\renewcommand{\thefootnote}{\fnsymbol{footnote}}
\footnotetext[2]{Corresponding author is Guanbin Li.}

\begin{abstract}
Active domain adaptation (ADA) aims to improve the model adaptation performance by incorporating active learning (AL) techniques to label a maximally-informative subset of target samples.
Conventional AL methods do not consider the existence of domain shift, and hence, fail to identify the truly valuable samples in the context of domain adaptation.
To accommodate active learning and domain adaption, the two naturally different tasks, in a collaborative framework, we advocate that a customized learning strategy for the target data is the key to the success of ADA solutions.
We present Divide-and-Adapt (DiaNA), a new ADA framework that partitions the target instances into four categories with stratified transferable properties.
With a novel data subdivision protocol based on uncertainty and domainness, DiaNA can accurately recognize the most gainful samples.
While sending the informative instances for annotation, DiaNA employs tailored learning strategies for the remaining categories.
Furthermore, we propose an informativeness score that unifies the data partitioning criteria.
This enables the use of a Gaussian mixture model (GMM) to automatically sample unlabeled data into the proposed four categories.
Thanks to the ``divide-and-adapt" spirit, DiaNA can handle data with large variations of domain gap.
In addition, we show that DiaNA can generalize to different domain adaptation settings, such as unsupervised domain adaptation (UDA), semi-supervised domain adaptation (SSDA), source-free domain adaptation (SFDA), etc. Project page: \url{https://github.com/Duojun-Huang/DiaNA-CVPR2023}
\end{abstract}
\section{Introduction}

Domain adaptation (DA) approaches strive to generalize model trained on a labeled source domain to a target domain with rare annotation~\cite{long2015dan,saito2019mme,ganin2016dann,li2021cdac} by coping with the domain disparity. Nevertheless, DA methods are significantly outperformed by their supervised counterparts due to the scarceness of annotation as demonstrated in~\cite{udasup1,udasup2,fu2021tqs}. In practice, it is cost-effective to get a moderate amount of target samples labeled to boost the performance of domain adaptation. Active learning (AL) approaches seek to select samples with uncertainty~\cite{lewis1994uncertain1,wang2016conf,huang2018entropy1,wang2014entropy2} and diversity~\cite{nguyen2004diverse2,sener2018coreset} to best benefit the model, which properly matches the demand. However, previous AL methods assume that both the labeled and unlabeled data follow the same distribution, such a strategy may become ineffective to the DA scenarios where the target data suffer from domain shift. The recently proposed active domain adaptation (ADA)~\cite{prabhu2021clue,fu2021tqs,xie2022eada} aims to resolve this issue by actively selecting the maximally-informative instances such that the performance of the transferred model can be best boosted with a limited annotation budget. 

\begin{figure}[t]
\centering
\includegraphics[width=80mm]{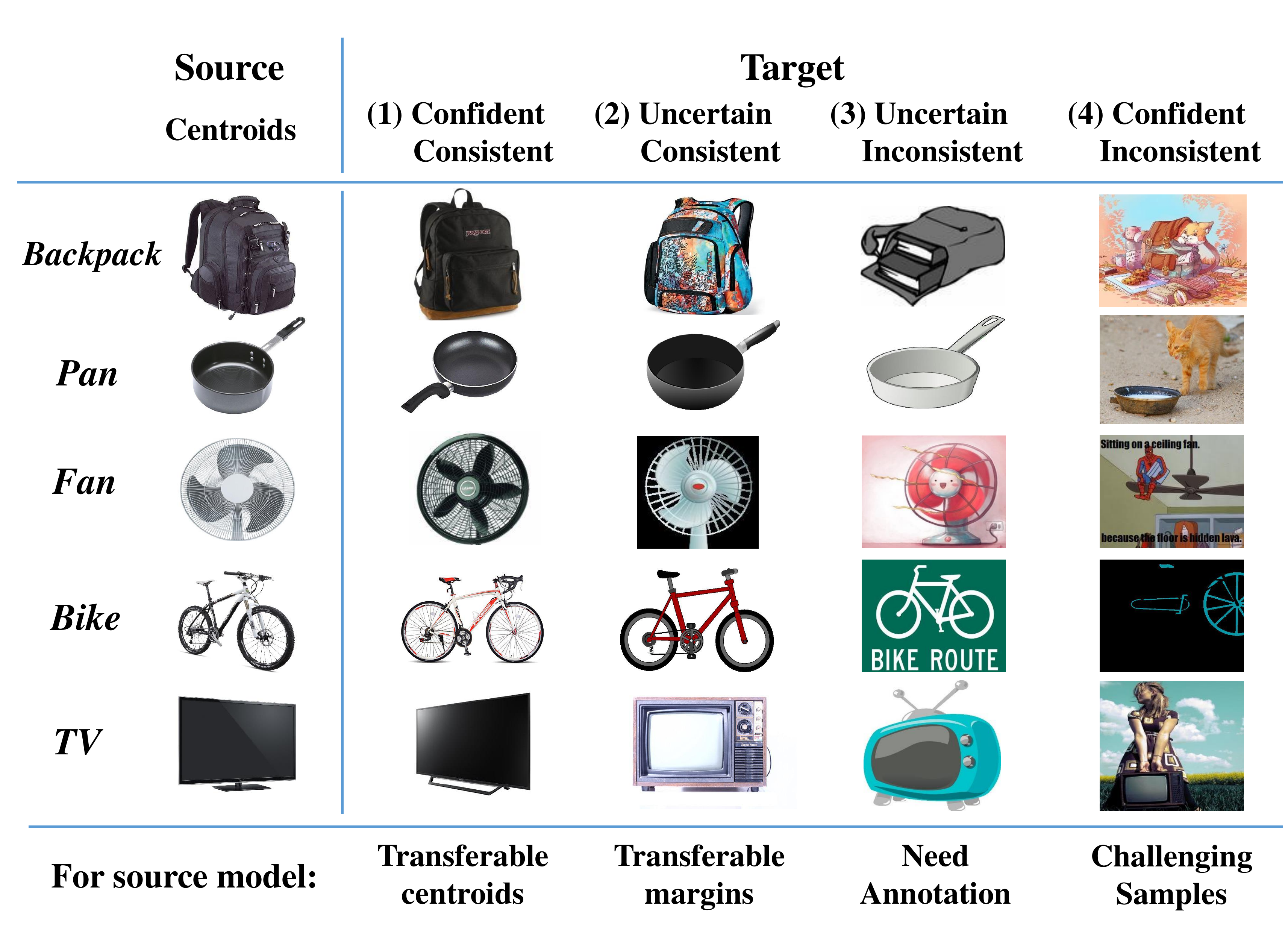}

\vspace{-0.3cm}
\caption{The illustration of our proposed Divide-and-Adapt mechanism to divide target samples into different data subsets for customized learning. 
}
\label{fig:GMM4_figs}
\vspace{-0.5cm}
\end{figure}

The key to the success of ADA is to strike a good balance between the highly coupled yet inherently different tasks: active learning and domain adaptation.
The real-world target data typically exhibit either of the two characteristics: \emph{source-like} or \emph{target-specific}. While the former has similar feature distribution with the source data, the latter tends to be the unique part of target domain and deviates greatly from the source distribution~\cite{su2020aada,fu2021tqs,zhang2022dacsfda}.
On one hand, to achieve domain adaptation, applying the same adaptation strategy to all target data equally cannot generalize well to scenarios with varying degrees of domain shift. This is particularly true when the gap between the source-like and target-specific data is unknown.
On the other hand, in active learning tasks, the samples with a learning difficulty will be more likely to be selected for labeling.
Nonetheless, with large domain gap, incorporating such difficult samples in the adaptation task would hamper the learning of the adaptation model, making the training highly unstable.
However, despite the impressive progress that has been made, none of the existing works has fully addressed the above issues.

In this work, we propose Divide-And-Adapt (\shortname{}), a novel ADA framework that can scale to large variations of domain gaps while achieving cost-effective data labeling with a significant performance boost for domain adaptation.
Our key observation is that customized learning strategies are vital for target data with different characteristics.
In particular, \shortname{} divides the target data into four subsets with different levels of transferable properties (see Figure~\ref{fig:GMM4_figs}), each of which is handled with a customized learning strategy.
Unlike traditional AL methods that would simply label the most uncertain data~\cite{wang2016conf,wang2014entropy2,yoo2019ll4al}, we propose to withhold the most challenging samples (Figure~\ref{fig:GMM4_figs} category (4)) for training the domain adaption models. 
Instead, the selected samples for active annotation would maintain a proper stimulus for the source model, providing informative domain knowledge without jeopardizing the training stability.  
The subdivision of target data is dynamically updated as the domain disparity is gradually mitigated with more labeled data.
Hence, the previous challenging samples could be classified as transferable in the later stage and exploited in the network training.

We introduce a novel protocol for subdividing the target samples for customized learning. In addition to the uncertainty of model prediction, we advocate that the consistency with the learned prior distribution, \emph{i.e.} the domainness, is another key criterion for active domain adaptation~\cite{su2020aada,fu2021tqs}.
To this end, we divide the target data into four categories as shown in Figure~\ref{fig:GMM4_figs} according to the domainness and uncertainty of the instances.
We further propose that the samples with 1) being \emph{uncertain} to the model and 2) having \emph{inconsistent} prediction with the label of its closest category prototype in the learned feature space (\emph{i.e.} high domainness) are the most ``profitable" instances for bringing informative knowledge of target domain if annotated.
Thereby, we identify the \emph{uncertain inconsistent} samples for labeling while applying tailored learning strategies for the remaining categories to boost the selectivity of the sampling. 

To avoid heuristic thresholding for data subdivision, we propose an automatic data sampling mechanism based on Gaussian mixture model (GMM). 
In particular, we propose an \emph{informativeness function} that incorporates the domainness and uncertainty in a unified scoring system.
The computed informativeness score of the labeled data is used to train a four-component GMM model, which is then applied to sample the unlabeled target data into four categories.

We evaluate \shortname{} over a large variety of domain shift scenarios on DomainNet~\cite{peng2019domainNet}, Office-Home~\cite{venkateswara2017officehome} and CIFAR-10~\cite{krizhevsky2009cifar10}. 
Furthermore, the proposed sampling strategy of \shortname{} can be generalized to various domain adaption problems with different supervision settings, including unsupervised domain adaptation (UDA), semi-supervised domain adaptation (SSDA), and source-free domain adaptation (SFDA). 

We summarize our contributions as follows:
\vspace{-2mm}
\begin{itemize}
    \item A general ``divide-and-adapt" framework, coded \shortname{}, for active domain adaptation that can handle diversified degrees of domain gaps while being able to generalize to different domain adaptation problems, including UDA, SSDA, and SFDA.
    \vspace{-2mm}
    \item A new target data partition strategy based on domainness and uncertainty that enables stratified learning to achieve more stable training, superior adaptation performance, and better generality. 
    \vspace{-2mm}
    \item A novel informativeness scoring system and the corresponding sampling paradigm based on GMM model for automatic data partitioning.
    \vspace{-2mm}
    \item New state-of-the-art performance over the mainstream public datasets in the task of active domain adaptation.   
\end{itemize}


\begin{figure*}
\centering

\includegraphics[width=14.5cm,height=5.5cm]{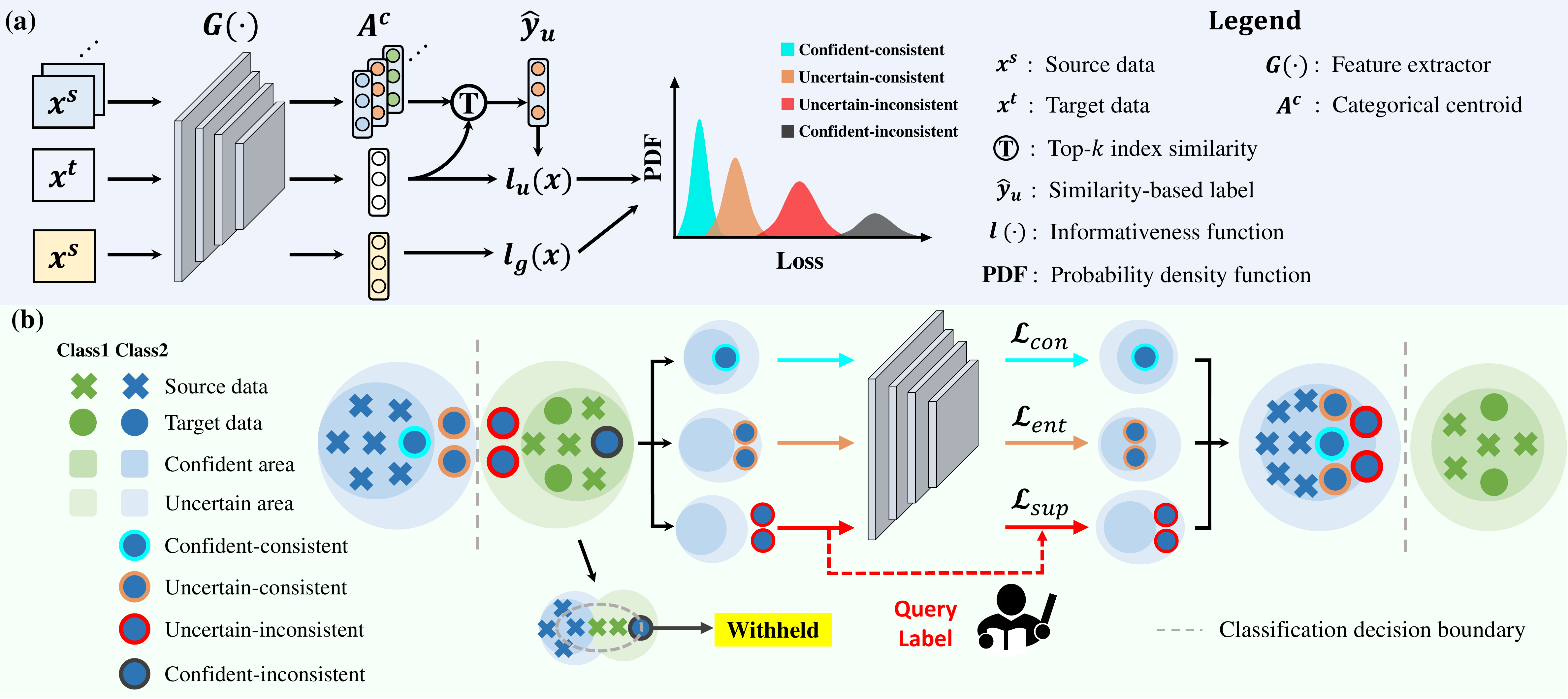}
\vspace{-0.3cm}
\caption{
We propose a general “divide-and-adapt” framework, named DiaNA, for active domain adaptation.
(a) A new target data partition strategy is presented to divide target data into four categories via building an informativeness scoring function that incorporates both domainness and uncertainty metrics. Besides, an automatic data partitioning function based on a learned four-component GMM model is applied to adaptively separate four sample categories from unlabeled target data. (b) To achieve customized learning, different auxiliary techniques are integrated into model fine-tuning w.r.t the four categories of unlabeled target data.
}
\vspace{-0.3cm}
\label{fig2:framework}
\end{figure*}

\section{Related Work}

\noindent
\textbf{Domain adaptation (DA).} Most of the prior methods involving domain adaptation follow the unsupervised domain adaptation (UDA) paradigm that only unlabeled data is accessible in the target domain.
Early UDA approaches~\cite{long2015dan,xie2016cmd,long2017jmmd} utilized maximum mean discrepancy (MMD) to minimize the discrepancy of features from different domains to address the domain shift, while recent adversarial learning based methods~\cite{ganin2016dann,long2018cada} have become popular solutions for UDA. Despite effectiveness, UDA is still significantly outperformed by its supervised variant as indicated in~\cite{udasup1,udasup2,fu2021tqs}. Semi-supervised domain adaptation (SSDA)~\cite{saito2019mme,li2021cdac,yan2022multi} allows a few target instances to be labeled and generally achieves improved performance. However, SSDA assumes that the labeled instances are passively provided in advance, depriving the choice of informative samples beneficial to the model. In this work, we propose an active domain adaption method to select the most informative samples, which can be incorporated into other DA methods to further boost their performance. 

\noindent
\textbf{Active Learning (AL).}
Active learning is proposed to select a maximally-informative subset of unlabeled data for annotation via a query function, in an effort to best improve the model performance with a limited annotation budget.
There exist several efforts to tackle the challenges involving active learning.
For example, the uncertainty-based AL methods mainly focus on designing query functions based on prediction confidence~\cite{wang2016conf}, entropy~\cite{huang2018entropy1,wang2014entropy2}, or margin~\cite{roth2006margin1,joshi2009margin2} of the posterior probabilities. On the other hand, the diversity-based approaches~\cite{guo2010diverse1,nguyen2004diverse2,sener2018coreset} concentrate on selecting samples that can well represent the entire dataset. Meanwhile, expected model change-based methods~\cite{freytag2014change1,liu2021isal} aim to query the instances that would lead to a significant change to the current model. Recently proposed active learning approaches attain better performance by utilizing hybrid strategies which taking multiple metrics into consideration~\cite{ash2019BADGE,parvaneh2022alpha}. Notwithstanding, the efficacy of previous AL resources may falter in DA scenarios as their selection strategy fails to meticulously consider the implications of the domain gap. 

\noindent
\textbf{Active Domain Adaptation (ADA).} 
Early ADA researches~\cite{su2020aada,fu2021tqs} propose to measure the uncertainty and domainness of each target instance via a domain discriminator with adversarial training. CLUE~\cite{prabhu2021clue} presents an entropy-weighted clustering algorithm to query uncertain and diverse samples in the target domain. Recently, SDM~\cite{xie2022sdm} is proposed to optimize a margin loss function for exploring target instances similar to potential hard samples in the source domain.
Nevertheless, the majority of existing ADA methods deliberately design hand-crafted query functions to evaluate sample's annotation value and apply the same adaptation strategy to all target data equally~\cite{xie2022eada,fu2021tqs,su2020aada}. 
The rigid criteria make them easily overfit to certain domain adaption scenarios, which limits the generalization of the method.
In contrast, we propose to integrate uncertainty and domainness metrics within a unified scoring function to select informative samples. Moreover, we apply tailored learning strategies for target data with different characteristics, constructing an effective framework capable of generalizing across large variations of domain shifts. 

\section{Divide-and-Adapt Framework}

\noindent \textbf{{Problem formulation.}}
Active domain adaptation aims at seeking an optimal model for a target domain when given labeled source data $\mathcal{S}=\{(x_{s}, y_{s})\}$ as well as unlabeled target data $\mathcal{U}=\{x_{u}\}$, assisted by the iteratively queried labeled data $\mathcal{T}=\{(x_{t}, y_{t})\}$ of the target domain.
Here, $\mathcal{S}$, $\mathcal{U}$, and $\mathcal{T}$ share the same label space over $C$ categories. Initially, $\mathcal{T}$ is an empty set.
The model is iteratively trained using $\mathcal{S}$, $\mathcal{T}$, and $\mathcal{U}$ for $R$ active learning loops until a given annotation budget $B$ is reached. In each active learning loop, $b= B/R $ samples from $\mathcal{U}$ would be selected, labeled by human experts, and then moved into $\mathcal{T}$.
The proposed Divide-and-Adapt framework consists of a feature extractor and a classifier.
Our goal is to design a query function to identify informative samples to annotate and utilize limited amount of labeled samples to best enhance the classification performance for the target domain. 

\noindent \textbf{Overview.}
In this paper, we propose Divide-And-Adapt, coded DiaNA, to tackle the problem of active domain adaptation. In detail, we first propose to design an informativeness function for subdividing the target samples into four categories, one of which is the informative target samples used for actively annotating.
Such a function is a unified scoring system that incorporates the domainness and the uncertainty.
To avoid heuristic thresholding for data subdivision, we propose an automatic data sampling mechanism to divide the unlabeled target data into four sample categories. This is achieved by learning a four-component Gaussian mixture model using semi-supervised Expectation-Maximization algorithm. Finally, we design tailored learning strategies for different categories of target samples during model fine-tuning, thus further improving the model performance. The overview of the proposed algorithm has been summarized in Figure~\ref{fig2:framework}.

\subsection{Informativeness Scoring Mechanism}

\begin{figure}[t]
\centering
\includegraphics[width=8cm,height=4.5cm]{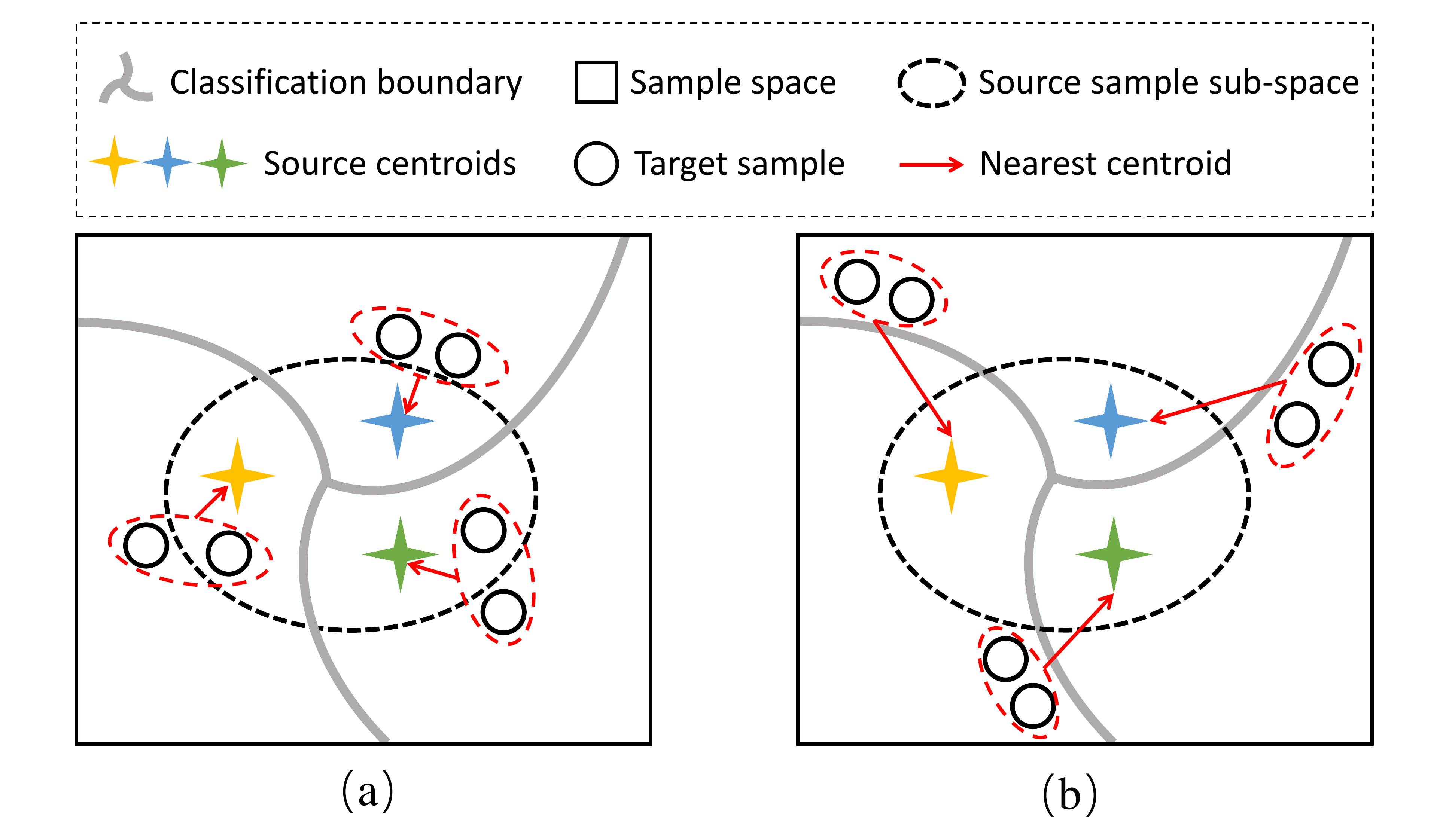}
\vspace{-0.5cm}
\caption{
(a) Source-like target samples are more likely to be close to the source centroid in the feature space with respect to its predicted category. (b) Target-specific target samples are prone to be inconsistent between model predictions and similarity-based labels. }
\vspace{-0.3cm}
\label{fig:consistency}
\end{figure}

Following previous ADA works~\cite{su2020aada,fu2021tqs,xie2022eada}, we aim to search for the target instances with high domainness and uncertainty. To achieve this goal, we propose to first estimate category-wise centroids based on the labeled samples, and then define a similarity-based label for each unlabeled target sample by computing its top-$k$ feature similarity with the categorical centroids. Finally, we integrate both the domainness and the uncertainty metrics into an informativeness function, which is formulated as the consistency between model prediction and similarity-based label. 

\paragraph{Categorical centroids.}
We propose to average the features of labeled samples to estimate the category-specific data distribution, i.e., category prototype.
Here, we formulate the centroid with respect to the $c$-th category as follows, 
\begin{equation}\label{class_centroid}
A^{c} = \frac{ \sum_{(x,y)\in \mathcal{S}} \mathds{1}{\{ y=c\}} \cdot G(x) }{\sum_{(x,y)\in \mathcal{S}} \mathds{1}\{ y=c\} }, 
\end{equation}
where $\mathds{1}\{\cdot\}$ is an indicator function, and $G(x)$ is a feature extractor that outputs the sample feature of $x$.
Note that, as the active learning loops proceed, labeled target samples would be incorporated into the calculation of categorical centroids to mitigate the impact of distribution change. 

\vspace{-3mm}
\paragraph{Similarity-based data label.}
As introduced in the works~\cite{han2020topkrank,li2021cdac}, samples share more top-$k$ indices in their respective lists of rank-ordered feature elements would have a higher probability of being the same category. 
To obtain the similarity-based label for each target instance, we first measure pairwise similarity of the indices of the sorted feature elements based on the magnitude. 
Then, we can formulate a similarity-based label for a sample $x$ from $\mathcal{U}$ as follows, 
\begin{equation}\label{sim_label}
\ddot{y}(x)=\mathop{\text{argmax}} \limits_{1 \le c \le C}\,\,\textbf{IoU}(\text{top}k(G(x)), \text{top}k(A^{c})),
\end{equation}
where $\text{top}k(\cdot)$ computes the top-$k$ feature element indices ranked according to their magnitudes. In addition,
$\textbf{IoU}(\cdot, \cdot)$ denotes an Intersection-over-Union function, which returns the extent of overlap of the two respective index lists.

We set the value of $k$ to be significantly smaller than the full dimension of the feature vector. In this way, the function $\text{top}k(\cdot)$ only extracts the principle components of the feature of sample. Therefore, the IoU function in Eq.~\ref{sim_label} is equivalent to measuring pairwise image sample similarity under a low-resolution condition.
In ADA tasks, the training of model is inevitably dominated by the source domain especially in the early stage~\cite{xie2022sdm}. Therefore, source-like target samples that are similar to the source domain would naturally have more accurate representation than the target-specific ones that are distinct from the source. Under the low-resolution condition, i.e., $k$ set to be small, source-like samples tend to have the same similarity-based label as the model predicted class thanks to their reliable and discriminative features extracted by the model. In contrast, target-specific samples are prone to produce inconsistent results since they are typically underfitted by the current model as shown in Figure~\ref{fig:consistency}. Based on the analysis above, we utilize the consistency of the model predicted class and similarity-based label to evaluate the domainness of each target sample. Detailed theoretical and experimental supports are demonstrated in our supplementary. 

\vspace{-3mm}
\paragraph{Informativeness function.}

To explore the most informative target samples, we integrate the domainness and uncertainty into a unified informativeness function (InfoF). The informativeness score for a sample $x$ from unlabeled target domain can be formulated as follows, 
\begin{equation}\label{l_ux}
\ell_{u}(x) = - \sum_{c=1}^C \mathds{1}\{c=\ddot{y}(x)\} \cdot \log P_{c}(x), 
\end{equation} 
where $P(x)$ denotes the probabilistic prediction of a sample $x$ from the network model and $P_{c}(x)$ is the $c$-th element of the vector $P(x)$.
According to Eq.~\ref{l_ux}, we can divide unlabeled target samples into two subsets depending on whether the model predicted class and the similarity-based label are identical: consistent and inconsistent subsets, while both sets have confident samples and uncertain samples depending on whether the predicted probability is greater than a given confidence threshold. We name these four categories of unlabeled target data as \textit{confident-consistent (CC)},  \textit{uncertain-consistent (UC)}, \textit{uncertain-inconsistent (UI)}  and \textit{confident-inconsistent (CI)}, respectively.
In this way, samples that are both uncertain to the model and contain target-specific knowledge can be extracted based on the InfoF score. In next section, we devise a strategy for informative sampling to adapatively distinguish those subsets from unlabeled target samples.

\begin{algorithm}[t]
\DontPrintSemicolon
   
    \KwInput{ $\mathcal{D}_{L}$; $\mathcal{D}_{U}$; $\alpha$ } 
        
   \KwOutput{Optimal GMM parameters $\mathcal{W}$}
    
    Initialize $\mathcal{W}\!=\!\{(\pi_{k}, \mu_{k}, \sigma_{k})|1\le k\le K \}$ using $\mathcal{D}_{L}$ via Bayes estimation.
    
    \While{$\mathcal{W}$ is not converged}{
    
    $\{{(\ell_g^i, p_g^i)}\}_{i=1}^{|\mathcal{D}_L|}\leftarrow\mathcal{D}_L$ 
    
    $\{{(\ell_u^j, )}\}_{j=1}^{|\mathcal{D}_U|}\leftarrow\mathcal{D}_U$ 
    
    \tcp{E step:}
    
    $\gamma_{gk}(\ell_g^i) = \mathds{1}\{p_g^i=k\},$
    
    $\gamma_{uk}(\ell_u^j) = \frac{\pi_{k} \mathcal{N}(\ell_u^j; {{{\mu}}_{k}},{{{\sigma}}_{k}})}{\sum_{k^{\prime}=1}^{K} \pi_{k^{\prime}}\mathcal{N}(\ell_u^j; {{{\mu}}_{k^{\prime}}},{{{\sigma}}_{k^{\prime}}})},$
    
    \tcp{M step:}
    
    ${\pi}_{k}=\frac{\alpha \sum_{i}\gamma_{gk}(\ell_g^{i}) + (1-\alpha)\sum_{j}\gamma_{uk}(\ell_u^{j}) }{\alpha \left\vert \mathcal{D}_{L} \right\vert + (1-\alpha)\left\vert \mathcal{D}_{U} \right\vert } $;
    
    ${\mu}_{k}=\frac{\alpha \sum_{i}\gamma_{gk}(\ell_g^{i})\cdot\ell_{g}^{i} + (1-\alpha)\sum_{j}\gamma_{uk}^{j}\cdot \ell_{u}^{j} }{\alpha \sum_{i}\gamma_{gk}(\ell_g^{i}) + (1-\alpha)\sum_{j}\gamma_{uk}(\ell_u^{j})} $;
    
    ${\sigma}_{k}=\frac{\alpha \sum_{i}\gamma_{gk}(\ell_g^{i})\cdot (\ell_{g}^{i}-\mu_{k})^{2} + (1-\alpha)\sum_{j}\gamma_{uk}\cdot(\ell_u^{j}) (\ell_u^{j}-\mu_{k})^{2}  }{\alpha \sum_{i}\gamma_{gk}(\ell_g^{i}) + (1-\alpha)\sum_{j}\gamma_{uk}(\ell_u^{j}) } $;
    
    }

\caption{The training procedure of four-component Gaussian Mixture Model (GMM) using semi-supervised  Expectation-Maximization}

\label{alg:algo1}
\end{algorithm}

\subsection{Informative Sampling Function}\label{sec:sec3.2}

During training, the varying scale of InfoF scores obtained above would hinder accurate data categorization using manual data division, leading to a biased sampling strategy and degraded performance.
Therefore, our work here formulates an Informative Sampling Function to achieve the goal of adaptively selecting informative target samples.

In general, it can be observed that the four categories of unlabeled target data as stated above are significantly separable in the distributions of the InfoF scores, making it possible to learn mixture models to separate four sample partitioning from unlabeled target data. Thereafter, by learning the distribution of InfoF scores using a four-component Gaussian mixture model (GMM), the target data can be adaptively mapped to the probability distribution of four categories and then be separated, thereby avoiding sampling bias that may be created by manual division.

In specific, we first take inspiration from~\cite{yan2017gaussian} that labeled samples can be used to construct supervision information, therefore building a more trustworthy GMM model with $K$ Gaussian components (namely $K=4$ in this work) via a semi-supervised  Expectation-Maximization algorithm. 
To this end,  we then calculate the InfoF score for a labeled sample with  $(x, y)$ from $\mathcal{S}$ or $\mathcal{T}$ as follows,
\begin{equation}\label{lab_loss}
\ell_{g}(x, y) = - \sum_{c=1}^C \mathds{1}{\{c=y\}} \cdot \log P_{c}(x).
\end{equation} 
The similarity-based label is determined using categorical centroids of the labeled data, resulting in the totally identical similarity-based label and model prediction for labeled samples. In order to obtain reliable supervision and build a better GMM model, we have replaced the similarity-based label with the ground-truth class label here.

Afterwards, to obtain the supervision information for training GMM, we here define a piecewise function to obtain the observation labels that assign all labeled samples across domains to the corresponding Gaussian components. 
Thus, the observation label of a candidate instance $x$ can be computed as,
\begin{equation}
\label{partition}
q(x, y)=\left\{
\begin{array}{rcl}
1, \quad &  { \max P(x) \geq \tau \; \text{and}\; y = \argmax P(x) },\\
2, \quad &  { \max P(x) < \tau \; \text{and}\; y = \argmax P(x) },\\
3, \quad &  { \max P(x) < \tau \; \text{and}\; y \ne \argmax P(x) },\\
4, \quad &  { \max P(x) \geq \tau \; \text{and}\; y \ne \argmax P(x) }, 
\end{array} \right.
\end{equation}
where the numbers, i.e., $k=1,\cdots,4$, indicate the indices of different Gaussian components corresponding to the {CC}, {UC}, {UI} and {CI} sample subsets, respectively. As well, $\tau$ is a confidence threshold to classify labeled samples into the confident subset and the uncertain subset. Furthermore, the comparison between the ground-truth and predicted class labels is used to divide them into the consistent subset and the inconsistent subset of labeled training samples. 

Once we obtain $\{\ell_g(x, y)|(x, y)\in\mathcal{S}\cup\mathcal{T}\}$, $\{\ell_u(x)|x\in\mathcal{U}\}$, and $q(x,y)$, 
we incorporate them into building training instances for learning the GMM model as follows,
\begin{equation}
\label{labelsplit}
\begin{split}
    \mathcal{D}_{L} \leftarrow \{(\ell_g(x, y), q(x, y))| (x, y)\in\mathcal{S}\cup\mathcal{T}\}, \\
\end{split}
\end{equation}
\begin{equation}
\begin{split}
    \mathcal{D}_{U} \leftarrow \{(\ell_u(x), )|x\in\mathcal{U}\}.
\end{split}
\end{equation}
For convenience, we denote the GMM model by $\mathcal{W}\!=\!\{(\pi_{k}, \mu_{k}, \sigma_{k})|1\le k\le K \}$. Afterwards, given a training instance $\ell_x$ from $\mathcal{D}_{L}$ or $\mathcal{D}_{U}$, the output of the probability density functions of the GMM model regarding $\ell_x$ can be computed as 
\begin{equation}\label{gmm}
p(\ell_x) = \sum_{k=1}^{K} \pi_{k} \mathcal{N}(\ell_x; \mu_{k}, \sigma_{k} ), 
\end{equation}
where $\pi_{k}$ represents the weight of the $k$-th Gaussian component subject to $\sum_{k=1}^K \pi_{k} = 1$ and $0 \le \pi_k \le 1$ for $k=1,\cdots, K$, while $\mu_{k}$ and $\sigma_{k}$ are vectors denoting the mean and variance of such a component. Here, $\mathcal{N}(\cdot; \mu_{k}, \sigma_{k} )$ is used to model the Gaussian distribution of the $k$-th component.
Note that a given coefficient $\alpha=|\mathcal{D}_{L}|/|\mathcal{D}_{U}|$ is used for the weighted integration of the labeled and unlabeled training instances. 
Thus, we can summarize the training procedure of the GMM model using a semi-supervised EM algorithm in Algorithm~\ref{alg:algo1} according to~\cite{yan2017gaussian}. 

After obtaining the trained GMM parameters $\mathcal{W}$, the probability of a sample $x$ from $\mathcal{U}$ being classified into the $k$-th component can be computed as follows,
\begin{equation}
\begin{split}
\text{Pr}(z=k|x, \mathcal{W}) = \frac{\pi_{k} \mathcal{N}(\ell_{u}(x); {{{\mu}}_{k}},{{{\sigma}}_{k}})}{\sum_{k^{\prime}=1}^{K} \pi_{k^{\prime}}\mathcal{N}(\ell_{u}(x); {{{\mu}}_{k^{\prime}}},{{{\sigma}}_{k^{\prime}}})}, 
\end{split}
\label{GMM_probs}
\end{equation} 
where $z$ is a discrete variable to infer which component the sample $x$ belongs to.

\subsection{Candidate Selection and Training Objectives}
After acquiring the informativeness sampling function, it is utilized for carrying out data partitioning. 
For ADA, the main focus should be on annotating samples that are uncertain to the model and can effectively represent the target dataset, which is termed as \textit{uncertain-inconsistent} samples within our subdivision framework. 
Therefore, during each sampling step, we first rank the unlabeled target samples in descending order according to the scores predicted by $\text{Pr}(z=3|\cdot, \mathcal{W})$, and then the top $b$ examples are selected to annotate and moved into labeled target data. Following previous ADA works, we apply the standard cross-entropy loss to the labeled data as follows: 
\begin{equation}\label{sup}
\mathcal{L}_{sup} = \mathop{\mathds{E}}\limits_{(x,y) \sim \mathcal{S}\cup\mathcal{T}} \big [- \sum_{c=1}^C \mathds{1}\{c=y\} \cdot \log P_{c}(x) \big ].
\end{equation}

To boost the selectivity of the sampling strategy, we further design tailored training techniques for the other data subsets with varied transferable properties. We perform data partitioning on the remaining unlabeled target samples, i.e., $\mathcal{U}\leftarrow\mathcal{U} \backslash \mathcal{T}$, according to the posterior probabilities regarding the rest of Gaussian components. 
\begin{equation}\label{gmm_i}
\hat{\mathcal{U}}_{k} =\{x| \argmax  \textbf{\text{Pr}}(x) = k, \forall (x, )\in {\mathcal{U}}\},  \text{for } k=1, 2, 4,
\end{equation} 
where $\hat{\mathcal{U}}_{1}$, $\hat{\mathcal{U}}_{2}$ and $\hat{\mathcal{U}}_{4}$ refer to $\hat{\mathcal{U}}_{\text{CC}}$, $\hat{\mathcal{U}}_{\text{UC}}$, $\hat{\mathcal{U}}_{\text{CI}}$. Here, $\textbf{\text{Pr}}(x)$ is the concatenated probability vector corresponding to different components, i.e., $\textbf{\text{Pr}}(x)=[\text{Pr}(z=1|x, \mathcal{W}), \text{Pr}(z=2|x, \mathcal{W}),\text{Pr}(z=3|x, \mathcal{W}), \text{Pr}(z=4|x, \mathcal{W})]$  w.r.t the sample $x$.

As shown in Figure~\ref{fig:GMM4_figs}(4), the source-oriented model could produce confident but unreliable predictions for some target samples having large domain gap with the source. Samples from $\hat{\mathcal{U}}_{\text{CI}}$ have extremely controversial results between model prediction and similarity-based label, which are most likely to be some excessively challenging instances. Therefore, we withhold them in the current stage to avoid jeopardizing the training stability. 

The samples in $\hat{\mathcal{U}}_{\text{CC}}$ have confident and reliable predictions, which tend to be some class centroid in the target feature space. Therefore, we apply consistency regularization to them to enforce the prediction consistency under different perturbations, which can be formulated as follows: 
\begin{equation}\label{l_con}
\mathcal{L}_{con} = \mathop{\mathds{E}}\limits_{(x, ) \sim \hat{\mathcal{U}}_{\text{CC}}}\big [- \sum_{c=1}^C \mathds{1}\{c=\ddot{y}(x)\} \cdot \log P_{c}(\text{\textbf{Aug}}(x)) \big ], 
\end{equation}
where $\text{\textbf{Aug}}(\cdot)$ is a function to create perturbations for the sample $x$ using classic data augmentation techniques such as AutoAugment~\cite{cubuk2019autoaugment}. 

For data group $\hat{\mathcal{U}}_{\text{UC}}$ whose prediction tend to be reliable but uncertain, we directly minimize the entropy of the predictions to  using the following conditional entropy loss: 
\begin{equation}\label{entropy_min}
\mathcal{L}_{ent} = \mathop{\mathds{E}}\limits_{(x, ) \sim \hat{\mathcal{U}}_{UC}} \big [ -\sum_{c=1}^C P_{c}(x) \log P_{c}(x) \big ]. 
\end{equation} 

In summary, the overall loss functions used to further optimize the model can be formulated as follows,
\begin{equation}\label{total_loss}
\mathcal{L} = \mathcal{L}_{sup} + \lambda_{c} \mathcal{L}_{con} + \lambda_{e} \mathcal{L}_{ent}
\end{equation} 
where $\lambda_{c}$ and $ \lambda_{e} $ are the weights to trade off different loss terms during the training process. As shown in Figure \ref{fig2:framework}, $\mathcal{L}_{sup}$ can calibrate the potentially underfitting of uncertain and target-specific samples. $\mathcal{L}_{con}$ can guide the model to learn global cross-domain clustering while $\mathcal{L}_{ent}$ 
serves to minimize the distance of data from the same class. As demonstrated in Sec.~\ref{Ablation}, these training objectives are complementary to each other to best adapt the model to the target domain. 

\subsection{Compatibility with UDA/SSDA/SFDA}
The proposed sampling strategy of DiaNA requires no additional network modules like a domain discriminator or multiple classifiers~\cite{su2020aada,fu2021tqs}, making it easy to be integrated into existing DA frameworks, such as unsupervised domain adaptation (UDA), semi-supervised domain adaptation (SSDA), and source-free domain adaptation (SFDA); see supplementary document for the implementation details. It should be noted that  we further design a variant of \shortname{} for the SFDA settings, due to the unavailability of source data. As evidenced by~Sec.~\ref{combination_exp}, when integrated into diverse DA techniques, the proposed sampling strategy can deliver
more performance improvement than random sample selection, thereby demonstrating its compatibility in various DA settings. To the best of our knowledge, we are the first to propose an active selection strategy that can generalize to UDA, SSDA, and SFDA methods.

\begin{table*}[t]
\centering
    \resizebox{\textwidth}{!}{
    \scriptsize
    \begin{tabular}{l|cccccccccccccccccc|ccc}
        \toprule
        \multirow{2}{*}{\centering Method} &\multicolumn{3}{c}{{R} $\to$ {C} } & \multicolumn{3}{c}{{C} $\to$ {S}} & \multicolumn{3}{c}{{S} $\to$ {P} } & \multicolumn{3}{c}{{C} $\to$ {Q}} & \multicolumn{3}{c}{{R} $\to$ {S} } & \multicolumn{3}{c}{{R} $\to$ {P} } & \multicolumn{3}{c}{AVG} \\
        & 1k & 2k& 5k & 1k & 2k& 5k & 1k & 2k& 5k & 1k & 2k & 5k & 1k & 2k & 5k & 1k & 2k& 5k & 1k & 2k& 5k \\ 
        \midrule            
        Random  & 50.9 & 55.2 & 61.5 & 42.6 & 44.8 & 49.4 & 41.0 & 44.4 & 50.1 & 26.3 & 33.2 & 43.0 & 38.2 & 40.7 & 46.9 & 46.5 & 48.8 & 53.7 & 40.9 & 44.5 & 50.8 \\ 
        CONF  & 48.6 & 54.2 & 61.3 & 41.9 & 43.6 & 48.0 & 41.6 & 44.2 & 48.5 & 25.5 & 32.7 & 42.8 & 36.9 & 39.4 & 45.3 & 47.1 & 47.8 & 51.2 & 40.3 & 43.6 & 49.5 \\ 
        Entropy~\cite{wang2014entropy2} & 48.0 & 54.2 & 61.9 & 42.7 & 45.2 & 48.9 & 41.1 & 43.7 & 48.7 & 24.6 & 31.0 & 42.1 & 37.0 & 40.4 & 45.7 & 46.5 & 48.4 & 52.2 & 40.0 & 43.8 & 49.9 \\ 
        Coreset~\cite{sener2018coreset} & 49.1 & 52.5 & 59.8 & 40.6 & 42.2 & 45.8 & 39.1 & 41.5 & 44.8 & 24.7 & 30.3 & 39.7 & 36.7 & 38.3 & 43.3 & 45.1 & 46.8 & 50.3 & 39.2 & 41.9 & 47.3 \\ 
        BADGE~\cite{ash2019BADGE} & 51.7 & 55.4 & 62.5 & 43.6 & 45.7 & 50.1 & 42.3 & 45.5 & 49.7 & 26.0 & 34.3 & 43.9 & 37.8 & 41.5 & 47.8 & 48.0 & 50.1 & 54.2 & 41.6 & 45.4 & 51.4 \\ 
        Alpha~\cite{parvaneh2022alpha} & 52.0 & 56.3 & 62.8 & 44.3 & 46.8 & 50.5 & 41.9 & 44.9 & 49.7 & 27.2 & 34.6 & 44.4 & 38.6 & 41.3 & 47.9 & 47.6 & 50.4 & 53.2 & 41.9 & 45.7 & 51.4 \\ 
        \midrule
        AADA~\cite{su2020aada} & 50.8 & 56.3 & 64.7 & 42.8 & 45.7 & 51.0 & 41.8 & 45.6 & 50.8 & 25.2 & 31.3 & 40.8 & 38.5 & 41.7 & 49.5 & 46.5 & 49.7 & 55.1 & 40.9 & 45.1 & 52.0  \\
        SDM~\cite{xie2022sdm} & 51.0 & 55.4 & 62.3 & 43.1 & 45.2 & 49.6 & 41.9 & 44.9 & 49.3 & 26.9 & 35.0 & 43.8 & 37.3 & 40.2 & 46.5 & 47.4 & 49.7 & 53.2 & 41.3 & 45.1 & 50.8  \\
        EADA~\cite{xie2022eada} & 51.5 & 56.1 & 63.9 & 44.2 & 46.3 & 51.7 & 43.8 & 46.1 & 50.3 & 27.5 & 34.8 & 42.9 & 39.0 & 41.5 & 46.4 & 48.6 & 50.4 & 54.1  & 42.4 & 45.9 & 51.2  \\
         CLUE~\cite{prabhu2021clue} & 54.1 & 59.4 & 66.0 & 45.3 & 48.7 & 53.7 & 44.0 & 48.0 & 53.2 & 28.8 & 35.5 & 43.2 & 40.4 & 44.4 & 51.1 & 50.0 & 52.7 & 56.4 & 43.8 & 48.1 & 53.9  \\
         \midrule
        {\shortname{}}(Ours) & \textbf{55.6} & \textbf{60.4} & \textbf{68.7} & \textbf{47.0} & \textbf{50.8} & \textbf{56.6} & \textbf{44.2} & \textbf{49.0} & \textbf{55.4} & \textbf{30.2} & \textbf{39.5} & \textbf{52.1} & \textbf{42.4} & \textbf{47.9} & \textbf{55.0} & \textbf{50.5} & \textbf{53.6} & \textbf{59.1} & \textbf{45.0} & \textbf{50.2} & \textbf{57.8} \\
        \bottomrule
        \end{tabular}
        }
        \vspace{-0.3cm}
        \caption{Comparison results (Accuracy: $\%$) on DomainNet with 1k, 2k and 5k labeling budgets. ``Random'' and ``CONF'' correspond to the classic AL approaches ``Random Sampling'' and ``Least-Confidence Sampling''.
        }\vspace{-3pt}
        \label{tab:domainnet}
\end{table*}

\begin{table*}[t]

    \centering
    \scriptsize    
        \begin{tabular}{l|cccccccccccc|cc}
        \toprule
        \multirow{2}{*}{Method} & \multicolumn{12}{c}{Office-Home}\\
        & A $\to$ C & A $\to$ P & A $\to$ R & C $\to$ A & C $\to$ P & C $\to$ R & P $\to$ A & P $\to$ C & P $\to$ R & R $\to$ A & R $\to$ C & R $\to$ P & AVG\\
        \midrule
        Random & 52.5 & 74.3 & 77.4 & 56.3 & 69.7 & 68.9 & 57.7 & 50.9 & 75.8 & 70.0 & 54.6 & 81.3 & 65.8  \\ 
        Entropy~\cite{wang2014entropy2} & 51.3 & 72.7 & 76.4 & 61.7 & 74.1 & 72.9 & 57.6 & 51.1 & 76.6 & 69.8 & 57.1 & 82.3 & 67.0  \\ 
        Coreset~\cite{sener2018coreset} & 52.8 & 73.3 & 75.5 & 59.8 & 73.4 & 70.8 & 58.1 & 52.6 & 75.8 & 69.3 & 56.5 & 82.5 & 66.7  \\ 
        BADGE~\cite{ash2019BADGE}  & 57.2 & 75.2 & 76.9 & 61.5 & 77.2 & 71.9 & 60.4 & 53.6 & 78.0 & 70.8 & 61.1 & 84.3 & 69.0  \\ 
        Alpha~\cite{parvaneh2022alpha}  & 57.0 & 79.4 & 78.2 & 61.6 & 78.0 & 74.1 & 58.9 & 54.2 & 78.1 & 71.7 & 61.1 & 84.6 & 69.7  \\ 
        \midrule
        AADA~\cite{su2020aada}  & 56.6 & 78.1 & 79.0 & 58.5 & 73.7 & 71.0 & 60.1 & 53.1 & 77.0 & 70.6 & 57.0 & 84.5 & 68.3\\    
        CLUE~\cite{prabhu2021clue}  & 58.0 & 79.1 & 77.0 & 61.3 & 78.0 & 73.1 & 60.4 & 55.9 & 77.9 & 70.9 & 60.3 & 84.1 & 69.7  \\ 
        TQS~\cite{fu2021tqs}  & 58.6 & 81.1 & 81.5 & 61.1 & 76.1 & 73.3 & 61.2 & 54.7 & 79.7 & 73.4 & 58.9 & 86.1 & 70.5\\
        SDM~\cite{xie2022sdm}  & 61.2 & 82.2 & 82.7 & 66.1 & 77.9 & 76.1 & 66.1 & 58.4 & 81.0 & 76.0 & 62.5 & 87.0 & 73.1 \\
        EADA~\cite{xie2022eada} & 63.6 & 84.4 & 83.5 & 70.7 & 83.7 & 80.5 & 73.0 & 63.5 & 85.2 & 78.4 & 65.4 & 88.6 & 76.7 \\

        \midrule
        {\shortname{}}(Ours) & \textbf{64.5} & \textbf{86.0} & \textbf{84.9} & \textbf{72.3} & \textbf{84.6} & \textbf{82.5} & \textbf{73.3} & \textbf{63.7} & \textbf{85.6} & \textbf{78.5} & \textbf{67.2} & \textbf{89.5} & \textbf{77.7} \\
        \bottomrule
        \end{tabular}
    \vspace{-0.3cm}
    \caption{ Comparison results (Accuracy: $\%$) on Office-Home with 5\% labeling budget.}
     \vspace{-0.5cm}
    \label{tab:officehome}
\end{table*}

\section{Experiments}\label{sec:sec-exp}
\subsection{Dataset}
We validate the effectiveness of the proposed approach {\shortname{}} on three standard benchmark datasets: DomainNet~\cite{peng2019domainNet}, Office-Home~\cite{venkateswara2017officehome} and CIFAR-10~\cite{krizhevsky2009cifar10}. On DomainNet, we select five domains namley Real (\textbf{R}), Clipart (\textbf{C}), Sketch (\textbf{S}), Painting (\textbf{P}), and Qucikdraw (\textbf{Q}), while we hire these domains to construct 6 adaptation scenarios. As in~\cite{fu2021tqs,xie2022sdm,xie2022eada}, we report 12 different adaptation scenarios on Office-Home, constructed from four domains: Real (\textbf{R}), Clipart (\textbf{C}), Art (\textbf{A}), and Product (\textbf{P}).  To further evaluate the generalization of the proposed sample selection strategy, we additionally report the classification results on CIFAR-10 with a standard active learning setting.

\subsection{Implementation}

Here, we mainly focus on the detailed descriptions of the settings involving active learning on DomainNet, Office-Home, and CIFAR-10, while more implementation details can be found in the supplementary material. In each adaptation scenario, we report the average accuracy over 3 trials.

\noindent \textbf{DomainNet.} Similar to {CLUE}~\cite{prabhu2021clue}, we consider ResNet-34~\cite{he2016resnet} as the network backbone, where such network model is pre-trained on all labeled source samples over 50 training epochs through standard supervision. In all adaptation cases, we set $B=5000$ and $R=10$. \\ 
\noindent \textbf{Office-Home.} To be fair, we follow the experimental settings of existing ADA works~\cite{fu2021tqs, xie2022sdm,xie2022eada} to conduct performance comparison. In particular, we also use ResNet-50~\cite{he2016resnet} as the backbone model, while we choose a $5\%$ proportion of unlabeled target data to set the labeling budget $B$, and $R$ is set to 5. \\
\noindent \textbf{CIFAR-10.} On this dataset, we select ResNet-34~\cite{he2016resnet} as the backbone model. We randomly choose  select 10\% training samples at random as the initially labeled data set, and we then annotate 4\% at each sampling step with a total annotation budget 30\%. \\
\vspace{-0.5cm}
\subsection{Main Results}
\vspace{-0.0cm}

\begin{figure*}[thbp]
  \centering
  \begin{subfigure}[b]{0.30\textwidth}
    \centering
    \includegraphics[width=0.8\linewidth]{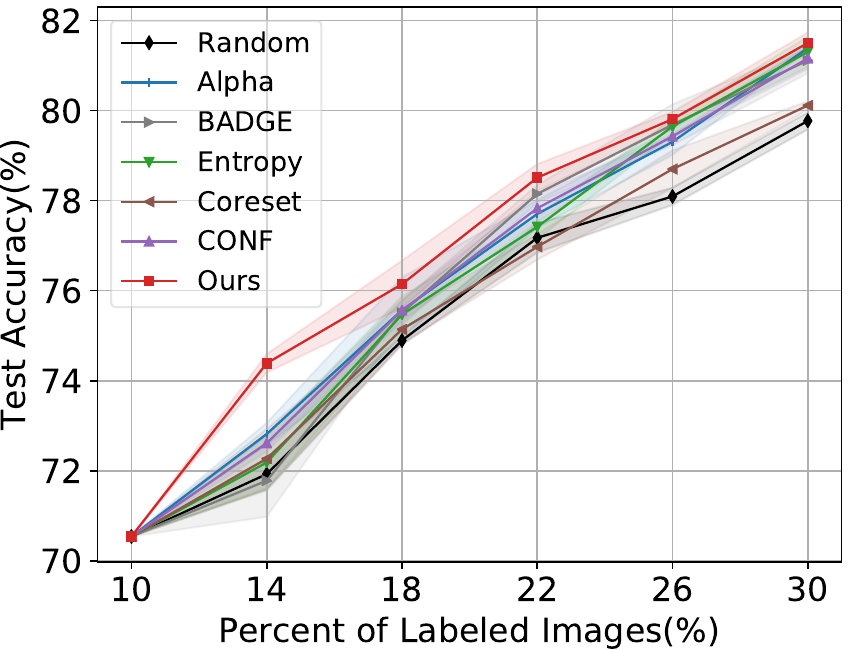}
    \caption[]%
    {{}}
    \label{fig:cifar10}
    \end{subfigure}\hspace{0.5em}
    \begin{subfigure}[b]{0.287\textwidth}
      \centering
      \includegraphics[width=0.8\linewidth]{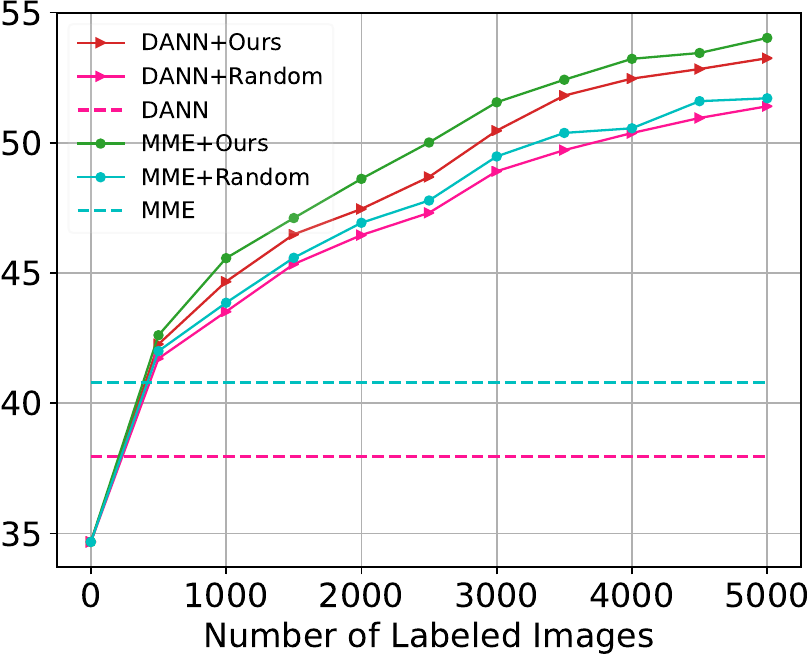}
      \caption[]%
      { }
      \label{fig:uda}
      \end{subfigure}\hspace{0.5em}
  \begin{subfigure}[b]{0.295\textwidth}
      \includegraphics[width=0.8\linewidth]{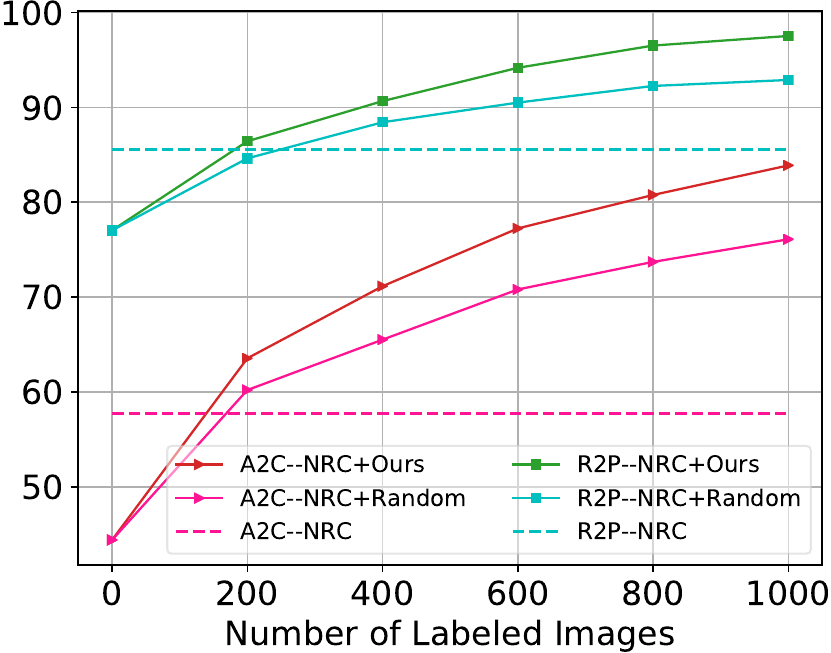}
      \caption[]%
      {{  }}
      \label{fig:sfda}
  \end{subfigure}\hspace{0.5em}
    \vspace{-0.3cm}
    \caption{(a) Comparison results on CIFAR-10. (b) Combination with UDA/SSDA methods. (c) Combination with SFDA methods.  } 
  
  \vspace{-0.5cm}
  \label{fig:other_exp}
\end{figure*}

\noindent\textbf{Results on DomainNet and Office-Home.}
We summarize the comparison with baselines in~Table~\ref{tab:domainnet} and Table~\ref{tab:officehome} on both datasets respectively. Note that we consider varying label budgets, namely i.e., 1k, 2k, and 5k, on DomainNet. We can see that {DiaNA} achieves the highest average accuracy in all cases on both datasets for solving the ADA tasks. Especially in the most difficult adaptation scenarios, e.g., C$\to$A and R$\to$C, on Office-Home, the proposed method can still exceed all the other methods by more than 1.6\% and 1.8\% respectively, demonstrating the effectiveness of DiaNA on the commonly used DA benchmark datasets.

\noindent\textbf{Results on CIFAR-10.} 
We also extend the proposed method to standard AL tasks on CIFAR-10, where the domain gap between labeled and unlabeled ones is minor. Figure \ref{fig:other_exp}\subref{fig:cifar10} lists the comparisons of {DiaNA} against classic AL strategies involving~\cite{sener2018coreset,wang2014entropy2,parvaneh2022alpha,ash2019BADGE}. It can be observed that compared to classic AL algorithms, with fewer labeling budgets, the effect of the proposed method is still noticeable, resulting from its significant performance improvement. This also indicates that 
our sampling strategy can be effectively applied to general active learning.

\noindent\textbf{Combination with UDA/SSDA/SFDA methods.} \label{combination_exp}
To validate the compatibility of the proposed {\shortname{}} with existing UDA, SSDA, and SFDA algorithms, we conduct the experiments for UDA and SSDA in the adaptation C$\to$S on DomainNet, and for SFDA in C$\to$A and R$\to$C on Office-Home. {DANN}~\cite{ganin2016dann}, {MME}~\cite{saito2019mme}, and {NRC}~\cite{yang2021NRC} correspond to the SOTA methods towards UDA, SSDA, and SFDA, respectively.
As shown in Figure~\ref{fig:other_exp}\subref{fig:uda} and Figure~\ref{fig:other_exp}\subref{fig:sfda}, combining the selection strategy of {\shortname{}} with other DA methods, i.e., {DANN+Ours}, {MME+Ours}, and {NRC+Ours}, significantly increases the performance of existing DA algorithms that select labeled target samples via random selection, namely {DANN+Random}, {MME+Random} and {NRC+Random}. This suggests that, adaptively considering both the domainness and uncertainty, {\shortname{}} is beneficial for the selection of informative target sample searching, while combined into diverse domain adaptation techniques.

\label{Ablation}
\begin{table}[t] 
    \centering
    \scriptsize 
    \resizebox{60mm}{!}{
    \begin{tabular}{l|cccc|c}
    \toprule[0.5pt]
    Method      & AL & $\mathcal{L}_{con}$ & $\mathcal{L}_{ent}$ & GMM & AVG \\
    \midrule[0.5pt]
    {\shortname{}}  & UI & $\checkmark$ & $\checkmark$ & $\checkmark$ & 77.7$\pm$0.2 \\
    {\shortname{}} w/o AL  & R & $\checkmark$ & $\checkmark$ & $\checkmark$ & 74.8$\pm$0.4 \\
    {\shortname{}} w/o $\mathcal{L}_{con}$ & UI & & $\checkmark$ & $\checkmark$ & 76.9$\pm$0.3 \\
    {\shortname{}} w/o $\mathcal{L}_{ent}$ & UI & $\checkmark$ & & $\checkmark$ & 75.8$\pm$0.3 \\
    {\shortname{}} w/o GMM & UI & $\checkmark$ & $\checkmark$ & & 74.0$\pm$0.2 \\
    \midrule[0.5pt]
    {\shortname{}}-UI  & UI   & & & $\checkmark$ & {72.1$\pm$0.4}  \\
    {\shortname{}}-UC & UC   &     &   & $\checkmark$   & 68.3$\pm$0.6 \\ 
    {\shortname{}}-CI & CI   &     &  & $\checkmark$    & 66.4$\pm$0.8 \\ 
    {\shortname{}}-CC & CC   &     &  & $\checkmark$    & 63.0$\pm$0.3 \\ 
    Random (Baseline) & R  &     &   &   & 65.8$\pm$0.5 \\ 
    \bottomrule[0.5pt]
    \end{tabular}
    } 
    \vspace{-0.3cm}
    \caption{Ablation study results of {\shortname{}}. The ``AL'' column indicates which type of data subset should be selected to be annotated during each sampling step. We evaluated the performance of all adaptation scenarios on Office-Home by averaging their accuracy.
    }
    \vspace{-0.5cm}
    \label{tab:ablation}
\end{table}

\subsection{Ablation study}

\noindent\textbf{Efficacy of the targeted training strategies.} 
To illustrate the effect of each component of the proposed {\shortname{}}, we conduct ablation study by removing the corresponding individual components. As displayed in Table~\ref{tab:ablation}, the proposed method exceeds all of the model variants by a large accuracy margin on average, demonstrating the effectiveness of each component.
By comparing our proposed entire approach with {\shortname{} w/o GMM}, manual data division strategy might considerably cause performance degradation, thereby in turn emphasizing the need for adaptive data subdivision protocol. 
Furthermore, as demonstrated, disrupting the correspondence between targeted data subsets provides empirical evidence supporting the validity of our proposed customized learning techniques; see supplementary document for more details.

\noindent\textbf{Efficacy of the sampling strategy.}
To provide insights into how effectively the domainness-based metric and the uncertainty-based metric work in the sampling step, we use different data subsets to replace the UI subset as active samples for annotation.
To eliminate the influence of $\mathcal{L}_{con}$ and $\mathcal{L}_{ent}$, we remove these two terms but only retain $\mathcal{L}_{sup}$ during training. As shown in Table~\ref{tab:ablation}, both {DiaNA-UC} and {DiaNA-CI} achieve better classification performance than baseline selection, thereby showing the effect of the proposed uncertainty metric and domainness metric. Moreover, active sampling on UI samples (DiaNA-UI) obtains the best performance among all the variants, further illustrating the superiority of the integration of both metrics. 
\section{Conclusions}
In this paper, we introduce a “divide-and-adapt” framework, named DiaNA, to address the problem of active domain adaptation. Specifically, we first design an informativeness function that jointly captures sample's uncertainty and domainness. Moreover, we devise an automatic data subdivision protocol to partition the target instances into four categories with different characteristics. While selecting the most informative samples to annotate, we also design tailored learning strategies for the other target data subsets with stratified transferable properties. Extensive experimental results, as well as ablation studies, have confirmed the superiority of the proposed approach.

\section*{Acknowledgments}
This work was supported in part by the Guangdong Basic and Applied Basic Research Foundation (NO.~2020B1515020048), in part by the National Natural Science Foundation of China (NO.~61976250), in part by the Shenzhen Science and Technology Program (NO.~JCYJ20220530141211024) and in part by the Fundamental Research Funds for the Central Universities under Grant 22lgqb25.

{\small
\bibliographystyle{ieee_fullname}
\bibliography{egbib}

\begin{thebibliography}{10}\itemsep=-1pt

\bibitem{arazo2019noisy1}
Eric Arazo, Diego Ortego, Paul Albert, Noel O’Connor, and Kevin McGuinness.
\newblock Unsupervised label noise modeling and loss correction.
\newblock In {\em International conference on machine learning}, pages
  312--321. PMLR, 2019.

\bibitem{ash2019BADGE}
Jordan~T Ash, Chicheng Zhang, Akshay Krishnamurthy, John Langford, and Alekh
  Agarwal.
\newblock Deep batch active learning by diverse, uncertain gradient lower
  bounds.
\newblock In {\em International Conference on Learning Representations}, 2019.

\bibitem{cubuk2019autoaugment}
Ekin~D Cubuk, Barret Zoph, Dandelion Mane, Vijay Vasudevan, and Quoc~V Le.
\newblock Autoaugment: Learning augmentation strategies from data.
\newblock In {\em Proceedings of the IEEE/CVF Conference on Computer Vision and
  Pattern Recognition}, pages 113--123, 2019.

\bibitem{freytag2014change1}
Alexander Freytag, Erik Rodner, and Joachim Denzler.
\newblock Selecting influential examples: Active learning with expected model
  output changes.
\newblock In {\em European conference on computer vision}, pages 562--577.
  Springer, 2014.

\bibitem{fu2021tqs}
Bo Fu, Zhangjie Cao, Jianmin Wang, and Mingsheng Long.
\newblock Transferable query selection for active domain adaptation.
\newblock In {\em Proceedings of the IEEE/CVF Conference on Computer Vision and
  Pattern Recognition}, pages 7272--7281, 2021.

\bibitem{ganin2016dann}
Yaroslav Ganin, Evgeniya Ustinova, Hana Ajakan, Pascal Germain, Hugo
  Larochelle, Fran{\c{c}}ois Laviolette, Mario Marchand, and Victor Lempitsky.
\newblock Domain-adversarial training of neural networks.
\newblock {\em The journal of machine learning research}, 17(1):2096--2030,
  2016.

\bibitem{guo2010diverse1}
Yuhong Guo.
\newblock Active instance sampling via matrix partition.
\newblock {\em Advances in Neural Information Processing Systems}, 23, 2010.

\bibitem{han2020topkrank}
Kai Han, Sylvestre-Alvise Rebuffi, Sebastien Ehrhardt, Andrea Vedaldi, and
  Andrew Zisserman.
\newblock Automatically discovering and learning new visual categories with
  ranking statistics.
\newblock In {\em International Conference on Learning Representations}, 2020.

\bibitem{he2016resnet}
Kaiming He, Xiangyu Zhang, Shaoqing Ren, and Jian Sun.
\newblock Deep residual learning for image recognition.
\newblock In {\em Proceedings of the IEEE conference on computer vision and
  pattern recognition}, pages 770--778, 2016.

\bibitem{huang2018entropy1}
Sheng-Jun Huang, Jia-Wei Zhao, and Zhao-Yang Liu.
\newblock Cost-effective training of deep cnns with active model adaptation.
\newblock In {\em Proceedings of the 24th ACM SIGKDD International Conference
  on Knowledge Discovery \& Data Mining}, pages 1580--1588, 2018.

\bibitem{joshi2009margin2}
Ajay~J Joshi, Fatih Porikli, and Nikolaos Papanikolopoulos.
\newblock Multi-class active learning for image classification.
\newblock In {\em 2009 ieee conference on computer vision and pattern
  recognition}, pages 2372--2379. IEEE, 2009.

\bibitem{kingma2014adam}
Diederik~P Kingma and Jimmy Ba.
\newblock Adam: A method for stochastic optimization.
\newblock {\em arXiv preprint arXiv:1412.6980}, 2014.

\bibitem{krizhevsky2009cifar10}
Alex Krizhevsky, Geoffrey Hinton, et~al.
\newblock Learning multiple layers of features from tiny images.
\newblock 2009.

\bibitem{lewis1994uncertain1}
David~D Lewis and Jason Catlett.
\newblock Heterogeneous uncertainty sampling for supervised learning.
\newblock In {\em Machine learning proceedings 1994}, pages 148--156. Elsevier,
  1994.

\bibitem{li2021cdac}
Jichang Li, Guanbin Li, Yemin Shi, and Yizhou Yu.
\newblock Cross-domain adaptive clustering for semi-supervised domain
  adaptation.
\newblock In {\em Proceedings of the IEEE/CVF Conference on Computer Vision and
  Pattern Recognition}, pages 2505--2514, 2021.

\bibitem{li2019dividemix}
Junnan Li, Richard Socher, and Steven~CH Hoi.
\newblock Dividemix: Learning with noisy labels as semi-supervised learning.
\newblock In {\em International Conference on Learning Representations}, 2019.

\bibitem{udasup2}
Jian Liang, Dapeng Hu, and Jiashi Feng.
\newblock Domain adaptation with auxiliary target domain-oriented classifier.
\newblock In {\em Proceedings of the IEEE/CVF Conference on Computer Vision and
  Pattern Recognition}, pages 16632--16642, 2021.

\bibitem{liu2021isal}
Zhuoming Liu, Hao Ding, Huaping Zhong, Weijia Li, Jifeng Dai, and Conghui He.
\newblock Influence selection for active learning.
\newblock In {\em Proceedings of the IEEE/CVF International Conference on
  Computer Vision}, pages 9274--9283, 2021.

\bibitem{long2015dan}
Mingsheng Long, Yue Cao, Jianmin Wang, and Michael Jordan.
\newblock Learning transferable features with deep adaptation networks.
\newblock In {\em International conference on machine learning}, pages 97--105.
  PMLR, 2015.

\bibitem{long2018cada}
Mingsheng Long, Zhangjie Cao, Jianmin Wang, and Michael~I Jordan.
\newblock Conditional adversarial domain adaptation.
\newblock {\em Advances in neural information processing systems}, 31, 2018.

\bibitem{long2017jmmd}
Mingsheng Long, Han Zhu, Jianmin Wang, and Michael~I Jordan.
\newblock Deep transfer learning with joint adaptation networks.
\newblock In {\em International conference on machine learning}, pages
  2208--2217. PMLR, 2017.

\bibitem{nguyen2004diverse2}
Hieu~T Nguyen and Arnold Smeulders.
\newblock Active learning using pre-clustering.
\newblock In {\em Proceedings of the twenty-first international conference on
  Machine learning}, page~79, 2004.

\bibitem{parvaneh2022alpha}
Amin Parvaneh, Ehsan Abbasnejad, Damien Teney, Gholamreza~Reza Haffari, Anton
  van~den Hengel, and Javen~Qinfeng Shi.
\newblock Active learning by feature mixing.
\newblock In {\em Proceedings of the IEEE/CVF Conference on Computer Vision and
  Pattern Recognition}, pages 12237--12246, 2022.

\bibitem{peng2019domainNet}
Xingchao Peng, Qinxun Bai, Xide Xia, Zijun Huang, Kate Saenko, and Bo Wang.
\newblock Moment matching for multi-source domain adaptation.
\newblock In {\em Proceedings of the IEEE/CVF international conference on
  computer vision}, pages 1406--1415, 2019.

\bibitem{prabhu2021clue}
Viraj Prabhu, Arjun Chandrasekaran, Kate Saenko, and Judy Hoffman.
\newblock Active domain adaptation via clustering uncertainty-weighted
  embeddings.
\newblock In {\em Proceedings of the IEEE/CVF International Conference on
  Computer Vision}, pages 8505--8514, 2021.

\bibitem{roth2006margin1}
Dan Roth and Kevin Small.
\newblock Margin-based active learning for structured output spaces.
\newblock In {\em European Conference on Machine Learning}, pages 413--424.
  Springer, 2006.

\bibitem{saito2019mme}
Kuniaki Saito, Donghyun Kim, Stan Sclaroff, Trevor Darrell, and Kate Saenko.
\newblock Semi-supervised domain adaptation via minimax entropy.
\newblock In {\em Proceedings of the IEEE/CVF International Conference on
  Computer Vision}, pages 8050--8058, 2019.

\bibitem{sener2018coreset}
Ozan Sener and Silvio Savarese.
\newblock Active learning for convolutional neural networks: A core-set
  approach.
\newblock In {\em International Conference on Learning Representations}, 2018.

\bibitem{su2020aada}
Jong-Chyi Su, Yi-Hsuan Tsai, Kihyuk Sohn, Buyu Liu, Subhransu Maji, and
  Manmohan Chandraker.
\newblock Active adversarial domain adaptation.
\newblock In {\em Proceedings of the IEEE/CVF Winter Conference on Applications
  of Computer Vision}, pages 739--748, 2020.

\bibitem{udasup1}
Yi-Hsuan Tsai, Wei-Chih Hung, Samuel Schulter, Kihyuk Sohn, Ming-Hsuan Yang,
  and Manmohan Chandraker.
\newblock Learning to adapt structured output space for semantic segmentation.
\newblock In {\em Proceedings of the IEEE conference on computer vision and
  pattern recognition}, pages 7472--7481, 2018.

\bibitem{venkateswara2017officehome}
Hemanth Venkateswara, Jose Eusebio, Shayok Chakraborty, and Sethuraman
  Panchanathan.
\newblock Deep hashing network for unsupervised domain adaptation.
\newblock In {\em Proceedings of the IEEE conference on computer vision and
  pattern recognition}, pages 5018--5027, 2017.

\bibitem{wang2014entropy2}
Dan Wang and Yi Shang.
\newblock A new active labeling method for deep learning.
\newblock In {\em 2014 International joint conference on neural networks
  (IJCNN)}, pages 112--119. IEEE, 2014.

\bibitem{wang2016conf}
Keze Wang, Dongyu Zhang, Ya Li, Ruimao Zhang, and Liang Lin.
\newblock Cost-effective active learning for deep image classification.
\newblock {\em IEEE Transactions on Circuits and Systems for Video Technology},
  27(12):2591--2600, 2016.

\bibitem{xie2022eada}
Binhui Xie, Longhui Yuan, Shuang Li, Chi~Harold Liu, Xinjing Cheng, and Guoren
  Wang.
\newblock Active learning for domain adaptation: An energy-based approach.
\newblock In {\em Proceedings of the AAAI Conference on Artificial
  Intelligence}, volume~36, pages 8708--8716, 2022.

\bibitem{xie2016cmd}
Junyuan Xie, Ross Girshick, and Ali Farhadi.
\newblock Unsupervised deep embedding for clustering analysis.
\newblock In {\em International conference on machine learning}, pages
  478--487. PMLR, 2016.

\bibitem{xie2022sdm}
Ming Xie, Yuxi Li, Yabiao Wang, Zekun Luo, Zhenye Gan, Zhongyi Sun, Mingmin
  Chi, Chengjie Wang, and Pei Wang.
\newblock Learning distinctive margin toward active domain adaptation.
\newblock In {\em Proceedings of the IEEE/CVF Conference on Computer Vision and
  Pattern Recognition}, pages 7993--8002, 2022.

\bibitem{yan2017gaussian}
Heng-Chao Yan, Jun-Hong Zhou, and Chee~Khiang Pang.
\newblock Gaussian mixture model using semisupervised learning for
  probabilistic fault diagnosis under new data categories.
\newblock {\em IEEE Transactions on Instrumentation and Measurement},
  66(4):723--733, 2017.

\bibitem{yan2022multi}
Zizheng Yan, Yushuang Wu, Guanbin Li, Yipeng Qin, Xiaoguang Han, and Shuguang
  Cui.
\newblock Multi-level consistency learning for semi-supervised domain
  adaptation.
\newblock {\em arXiv preprint arXiv:2205.04066}, 2022.

\bibitem{yang2021NRC}
Shiqi Yang, Joost van~de Weijer, Luis Herranz, Shangling Jui, et~al.
\newblock Exploiting the intrinsic neighborhood structure for source-free
  domain adaptation.
\newblock {\em Advances in Neural Information Processing Systems},
  34:29393--29405, 2021.

\bibitem{yoo2019ll4al}
Donggeun Yoo and In~So Kweon.
\newblock Learning loss for active learning.
\newblock In {\em Proceedings of the IEEE/CVF conference on computer vision and
  pattern recognition}, pages 93--102, 2019.

\bibitem{zeiler2012adadelta}
Matthew~D Zeiler.
\newblock Adadelta: an adaptive learning rate method.
\newblock {\em arXiv preprint arXiv:1212.5701}, 2012.

\bibitem{zhang2022dacsfda}
Ziyi Zhang, Weikai Chen, Hui Cheng, Zhen Li, Siyuan Li, Liang Lin, and Guanbin
  Li.
\newblock Divide and contrast: Source-free domain adaptation via adaptive
  contrastive learning.
\newblock {\em arXiv preprint arXiv:2211.06612}, 2022.

\end{thebibliography}
}

\newpage
\appendix
\section*{Supplementary Materials}


\newcommand\DoToC{%
  \startcontents
  \printcontents{}{2}{\textbf{Contents}\vskip3pt\hrule\vskip5pt}
  \vskip3pt\hrule\vskip5pt
}

\noindent
We present additional implementation details and analysis of our proposed method DiaNA in this supplementary material. 

\section{Experiment Details}

\paragraph{Implementation details.} 
We implement all the experiments using PyTorch\footnote{https://pytorch.org/}. 
For a fair comparison, most of experimental implementations involving both model training and active learning are consistent with the previous ADA works~\cite{fu2021tqs,xie2022sdm,prabhu2021clue}. The model is initially trained with labeled source data before the sampling steps. During the training phase, we train the model using an Adam~\cite{kingma2014adam} optimizer with the learning rate of 10$^{-5}$ for DomainNet and an AdaDelta~\cite{zeiler2012adadelta} optimizer with the learning rate of 0.1 for Office-Home, while the mini-batch size is set to 64 and 32 for DomainNet and Office-Home, respectively. 
Furthermore, we set the confidence threshold $\tau$ to 0.95, while we set $k$ used in pairwise feature similarity to 32 and 64 for ResNet-34 and ResNet-50, respectively. Finally, the loss weights $\lambda_{c}$ and $\lambda_{e}$ are set to 0.5 and 0.1 respectively. 

\paragraph{Combination with UDA/SSDA/SFDA.}
To illustrate the compatibility of the proposed DiaNA with existing UDA/SSDA/SFDA algorithms, we have shown the comparison results in the main text. Here, we give full details on the re-implementations of how to combine DiaNA with these DA methods. 
Specifically, we initially train the task model with all labeled source samples from $\mathcal{S}$ through supervision. 
Then, the targeted active samples from the unlabeled target data subset $\mathcal{U}$ would be selected by the proposed sampling strategy, annotated by the human experts, and then moved to the labeled target data subset $\mathcal{T}$.
Afterwards, in response to the ADA setup, we conduct domain alignment on $\mathcal{S}$ and $\mathcal{U}$ based on existing DA framework while simultaneously imposing a supervised loss on $\mathcal{T}$ to further optimize the model.

\begin{figure*}[thbp]
  \centering
  \begin{subfigure}[b]{0.45\textwidth}
    \centering
    \includegraphics[width=1.0\linewidth]{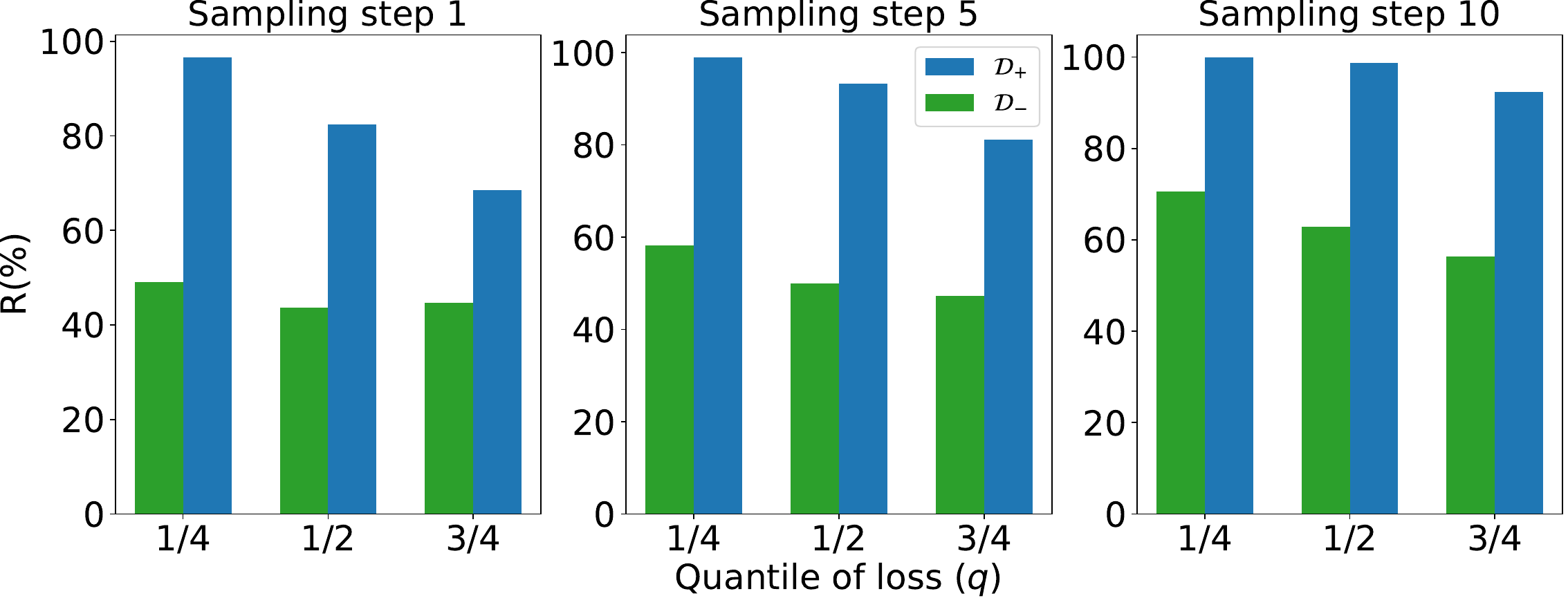}
    \caption[]%
    { DomainNet }
    \label{fig:dn_cr}
  \end{subfigure}\hspace{0.5em}
  \begin{subfigure}[b]{0.45\textwidth}
      \centering
      \includegraphics[width=1.0\linewidth]{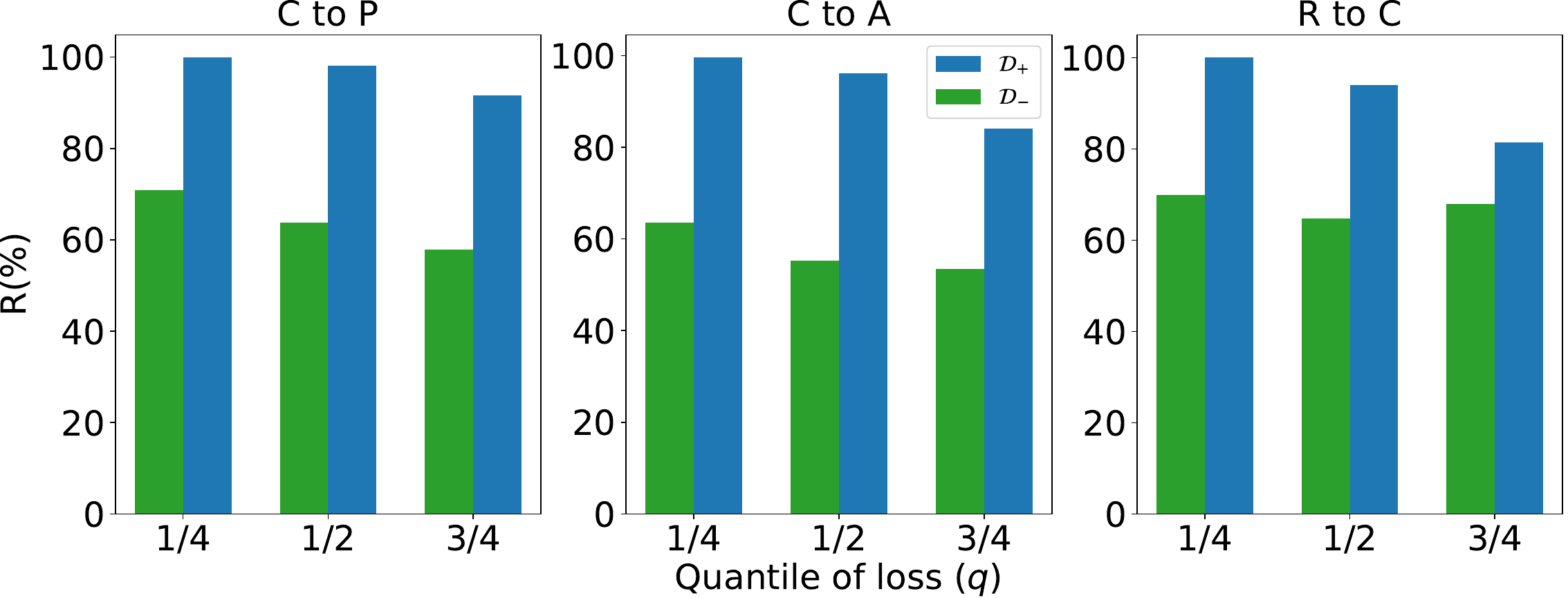}
      \caption[]%
      {Office-Home }
      \label{fig:oh_cr}
  \end{subfigure}\hspace{0.5em}
  \vspace{-5pt}
    \caption{Consistency rates of the well-learnt data subset $\mathcal{D}_{+}$ and the underfitted data subset $\mathcal{D}_{-}$ under diverse applications. The hyper-parameter $k$ is set to 32 for ResNet-34 on DomainNet and 64 for ResNet-50 on Office-Home. (a) Across different sampling steps in the adaptation scenario C$\to$S on DomainNet. (b) Across different adaptation scenarios in the first sampling step on Office-Home. 
    } 
  \vspace{-6pt}
  \label{fig:topk}
\end{figure*}

In this work, we need to construct training instances with the aid of labeled samples to obtain the supervision information for training GMM. However, in the SFDA scenario, the labeled source samples cannot be obtained again. Without the aid of labeled examples, this makes it temporarily impossible to divide the target samples into four categories during the first sampling step. Here, we propose to build a model variant of DiaNA to achieve the goal of training the GMM model. Specifically, we first searches for \textit{inconsistent} samples from unlabeled target data and then divide an \textit{uncertain} subset from inconsistent samples by combining the predicted confidence from the model.

Firstly, we select the target samples with high model prediction confidence as the proxy of the labeled data, which can be formulated as follows, 
\begin{equation}\label{L_hat}
\hat{\mathcal{L}} =\{(x,\hat{y})| \text{max } P(x) \ge t_{v}, \forall (x, )\in {\mathcal{U}}\}, 
\end{equation} 
where $\hat{y}={\text{arg max }} P(x)$ denotes the predicted class label of the sample $x$ by the model. The confidence threshold value is denoted as $t_{v}$ to screen confident samples from the target samples. Thanks to the relatively reliable prediction of confident samples, the sample-label pair $(x, \hat{y})$ can be utilized to build the substitutes of the labeled data. As it is necessary to calculate the categorical centroid for each category, the value of $t_{v}$ is set to 0.95 initially and will be iteratively reduced by 0.1 untill $\hat{\mathcal{L}}$ contains all the categories. After $\hat{\mathcal{L}}$ is obtained, the per-class categorical centroid $A^{c}$ can be estimated through Eq. (\textcolor{red}{1}) with $\mathcal{S}$ replaced by $\hat{\mathcal{L}}$. 
Further, the similarity-based label $\ddot{y}(x)$ for each unlabeled target sample $x$ is calculated through Eq. (\textcolor{red}{2}) in the main text. 
To obtain the samples with consistent model prediction and similarity-based label and uncertain model prediction (``uncertain-inconsistent'') as introduced in the main paper, we obtain the active samples in each sampling step as follows,
\begin{equation}\label{gmm_i_supp}
X_{3} =\{(x,\hat{y})|  \hat{y} \ne \ddot{y}(x) \, \textbf{and} \, \text{max} P(x) \le t_{c}, \forall (x, )\in {\mathcal{U}}\}, 
\end{equation} 
where $t_{c}$ denotes the confidence threshold value to select uncertain samples. The value of $t_{c}$ is set to $\frac{1}{C}+10^{-5}$ initially, where $C$ is the number of categories. Then $t_{c}$ will be iteratively increased by 0.1 untill $|X_{3}|$ reaches the annotation quota $b$ in each sampling step. 
For the following sampling steps, the labeled data will be obtained with $\hat{\mathcal{L}}$ if $A^{c}$ can not contain all the categories. Otherwise it will be obtained with $\mathcal{T}$ as the formulation Eq. (\textcolor{red}{1}).

\section{Further analysis of \shortname{}}

\begin{table*}[htbp]
    \centering
    
    \resizebox{0.76\textwidth}{!}{$
    \begin{tabular}{l|c|ccc}
    \toprule
         Dataset & Office-Home & \multicolumn{3}{c}{DomainNet}   \\
         Labeling Budget & 5\% & 1k & 2k & 5k   \\ \midrule
         1. DiaNA-$\mathcal{L}_{con}$/$\mathcal{L}_{ent}$ for {CI} & 68.2 / 74.6 & 44.3 / 44.6 & 49.5 / 49.9 & 56.8 / 57.6   \\ 
        2. DiaNA-$\mathcal{L}_{con}$ for {UC}/{UI}/{CI} & 73.8 / 73.0 / 72.0 & 44.2 / 43.3 / 42.6 & 49.0 / 48.2 / 47.4 & 55.4 / 55.7 / 54.7     \\ 
        3. DiaNA-$\mathcal{L}_{ent}$ for {CC}/{UI}/{CI} & 74.4 / 74.1 / 73.4 & 44.1 / 44.4 / 44.3 & 47.9 / 48.0 / 47.9 & 54.2 / 53.7 / 53.5    \\ 
        4. DiaNA  & \textbf{77.7} & \textbf{45.0} & \textbf{50.2} & \textbf{57.8}  \\ 
    \bottomrule
    \end{tabular}  
    $}
    \caption{Ablation study of the proposed customized strategy. The performance is evaluated by averaging the accuracy ($\%$) of all the adaptation scenarios. }
    \label{tab:more_ablation}
\end{table*}

\subsection{Analysis of the top-$k$ similarity}
\label{sec:topk}

\begin{table*}[htbp]
    \centering

    \resizebox{0.90\textwidth}{!}{$
        \begin{tabular}{l|cccccccccccc|cc}
        \toprule
        Method &  A $\to$ C & A $\to$ P & A $\to$ R & C $\to$ A & C $\to$ P & C $\to$ R & P $\to$ A & P $\to$ C & P $\to$ R & R $\to$ A & R $\to$ C & R $\to$ P & AVG\\
        \midrule
        Random & 57.7 & 29.3 & 28.6 & 40.8 & 37.3 & 34.1 & 38.4 & 63.6 & 34.5 & 28.0 & 55.9 & 24.9 & 39.4  \\ 
        BADGE~\cite{ash2019BADGE}& 63.6 & 49.3 & 46.8 & 64.0 & 50.7 & 50.0 & 71.2 & 72.7 & 51.8 & 68.8 & 72.7 & 53.3 & 59.6  \\ 
        CLUE~\cite{prabhu2021clue} & 69.5 & 52.9 & 50.0 & 60.8 & 54.7 & 50.5 & 56.0 & 67.3 & 57.7 & 56.8 & 71.8 & 49.3 & 58.1  \\ 

        \midrule
        {\shortname{}}(Ours) & \textbf{78.2} & \textbf{60.9} & \textbf{60.5} & \textbf{73.6} & \textbf{71.6} & \textbf{66.4} & \textbf{73.6} & \textbf{75.5} & \textbf{62.7} & \textbf{74.4} & \textbf{74.1} & \textbf{60.9} & \textbf{69.4} \\
        \bottomrule
        \end{tabular}
    $}
    \caption{ The error rate of the selected samples averaged over all the sampling steps. The experiment is conducted on Office-Home with 5\% labeling budget. }
    \vspace{-0.5cm}
    \label{tab:error_rate}
\end{table*}

Using categorical centroids and top-$k$ feature similarities, as stated in the main paper, we construct a domainness-based metric to distinguish between source-like samples and target-specific samples from unlabeled target data. 
In the context of active domain adaptation (ADA), the trained model is inherently biased towards a prominent region of the source domain with high data density~\cite{prabhu2021clue,xie2022sdm}, as the vast majority of sample labels come from the source. Hence, the source-like target samples, with high feature similarity to the source data, tend to be well learned by the model. On the contrary, as the unique part of target data distribution, the target-specific samples are more likely to suffer from underfitting by the model. Therefore, we aim to construct the domainness metric based on the prediction reliability of sample.

When the value of $k$ in top-$k$ similarity is set to be the full dimension of the feature vector and significantly small, the IoU function in the formulation of similarity based label is equivalent to measuring pairwise image sample similarity under full-resolution and low-resolution conditions, respectively. In the former case, almost all the samples would have identical model predicted class and the label of its closest category prototype in the feature space. When $k$ is set to be significantly small, only the samples with accurate and discriminative features extracted by the model can maintain the consistency between the labels of the two views, since the similarity label is obtained based on only the top-$k$ main components of the sample feature.
Here, we make an assumption that when $k$ is set to be small, the well-learned samples tend to maintain a consistent identity for these two labels thanks to their reliable and discriminative features extracted by the model. In contrast, the underfitted samples are more likely to produce a similarity-based label inconsistent with the class label predicted by the model. As the training of model is inevitably dominated by the source domain in the context of ADA, the source-like and target-specific samples in the target well correspond with the well-learned and underfitted samples. Therefore, we utilize the consistency of the model predicted class and similarity-based label to evaluate the domainness of each target sample.

Here, we conduct a validation experiment to investigate the feasibility of our assumption. 
According to~\cite{arazo2019noisy1, li2019dividemix}, given a trained model, the cross-entropy loss could illustrate how well the model fits the training examples. Hence, we hire it to divide the target samples into the well-learnt and underfitted subsets.
Specifically, we should first denote a function to evaluate the loss function value of a target sample $(x, )\in\mathcal{U}$ as follows, 
\begin{equation}\label{lx}
\ell(x) = - \sum_{c=1}^C \mathds{1}{\{c=y(x)\}} \cdot \log P_{c}(x),
\end{equation} 
where $y(x)$ denotes the actual label of such a sample $x$. After that, we can separate a well-learnt subset and an underfitted subset from all unlabeled target samples as follows, 
\begin{equation}\label{dw}
\mathcal{D}_{+} = \{x | \ell(x) \le \ell_{q}, \forall (x, )\in\mathcal{U} \},
\end{equation} 
\begin{equation}\label{du}
\mathcal{D}_{-} = \{x | \ell(x) > \ell_{q}, \forall (x, )\in\mathcal{U} \},
\end{equation} 
where $\ell_{q}$ is the quantile point of the sorted loss function values. In our implementation, we utilize the 1/4, 1/2, and 3/4 quantile points. Therefore, we define the consistency rate of each data subset as follows,
\begin{equation}\label{cr}
R  =  \frac{ \sum_{x\in \mathcal{D}_{*}} \mathds{1}{\{ \hat{y}(x)=\ddot{y}(x)\}} }{|\mathcal{D}_{*}|}, 
\end{equation} 
where $\mathcal{D}_{*}$ is a placeholder that represents $\mathcal{D}_{+} $ or $\mathcal{D}_{-} $. As shown in Figure \ref{fig:topk}, when $k$ is set to a value significantly smaller than the feature dimensions, namely 256 for ResNet-34 (or, 2048 for ResNet-50), the consistency rate of the well-learnt data subset is substantially higher than that of the underfitted counterpart. These results demonstrate the feasibility of the proposed assumption to distinguish between the well-learnt source-like samples and the underfitted target-specific samples.

\subsection{Analysis of customized learning strategy}
 
We conduct more ablation studies to verify the reasonability of our proposed customized learning strategy. For the tailored training objectives designed for different targeted data subsets, we replace the constrained data subset of $\mathcal{L}_{con}$/$\mathcal{L}_{ent}$ with incongruous subsets. As displayed in Table~\ref{tab:more_ablation} \#2-3, DiaNA significantly outperforms all of its variants, demonstrating the superiority of the customized training strategies. 
In addition, we also withhold CI samples that are incompatible with the current model due to their potential large domain gap with the source domain (see Figure.\textcolor{red}{1}(4) in main text). If CI samples are used as constraints, their significant discrepancies between the model's predictions and the similarity labels would jeopardize the training stability. Furthermore, Table~\ref{tab:more_ablation} \#1 reveals the performance of both datasets decreases as a result of additionally adding $\mathcal{L}_{con}$/$\mathcal{L}_{ent}$ to CI samples.

\subsection{Analysis of the Informative Sampling Function}
\begin{figure}[th]
\centering
\includegraphics[width=50mm]{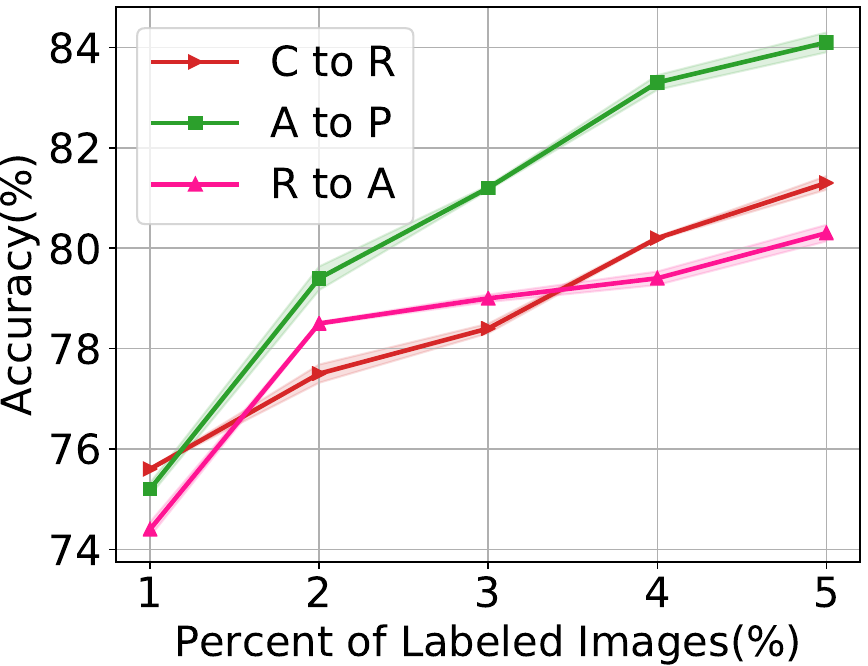}
\caption{Accuracy of identifying the four categories of unlabeled target data. }
\label{fig:gmm_acc}
\end{figure}

The data partitioning result produced by the Informative Sampling Function provides a fundamental support for identifying the most informative samples in $\mathcal{U}$. We extensively verify the efficacy of the proposed sampling function constructed based on the Gaussian Mixture Model (GMM). As stated in the main text, the predicted category of each unlabeled target sample in $\mathcal{U}$ is determined by the posterior probabilities of GMM.
We further obtain the four-category label according to the piecewise function described in Sec. \textcolor{red}{3.2} of the main text. The accuracy is calculated as the ratio of the correctly-classified samples in $\mathcal{U}$. It can be observed from Figure~\ref{fig:gmm_acc} that the sampling function is able to identify target samples belonging to the four categories. It should be noted that the accuracy of the model generally increases with the number of labeled images, indicating that the selection and adaptation are complementary to each other for achieving the best domain adaptation performance.

\subsection{Error rate of the selected samples}
The sampling strategy of the proposed \shortname{} aims to select the target-specific target samples for annotation, which are typically underfitted by the model as mentioned in Sec.~\ref{sec:topk}. To investigate the characteristics of the selected target samples, we report the error rates of all active samples selected by different strategies involving active learning and active domain adaptation. As shown in Table~\ref{tab:error_rate}, the error rates of \shortname{} are consistently higher than the other methods in all cases on Office-Home. This demonstrates that \shortname{} is capable of selecting 
these relatively hard-to-learn target samples. As illustrated in Figure \textcolor{red}{2(b)} in the main text, annotating these samples and applying supervision to them can potentially correct the model predictions for better adapting to the target data distribution.

\subsection{Hyper-parameter sensitivity. }

\begin{figure}[t]
\centering
    \hspace{-3mm}
    \subfloat[Analysis of $k$]{
    \label{fig:sense-m}
    \includegraphics[width=0.233\textwidth]{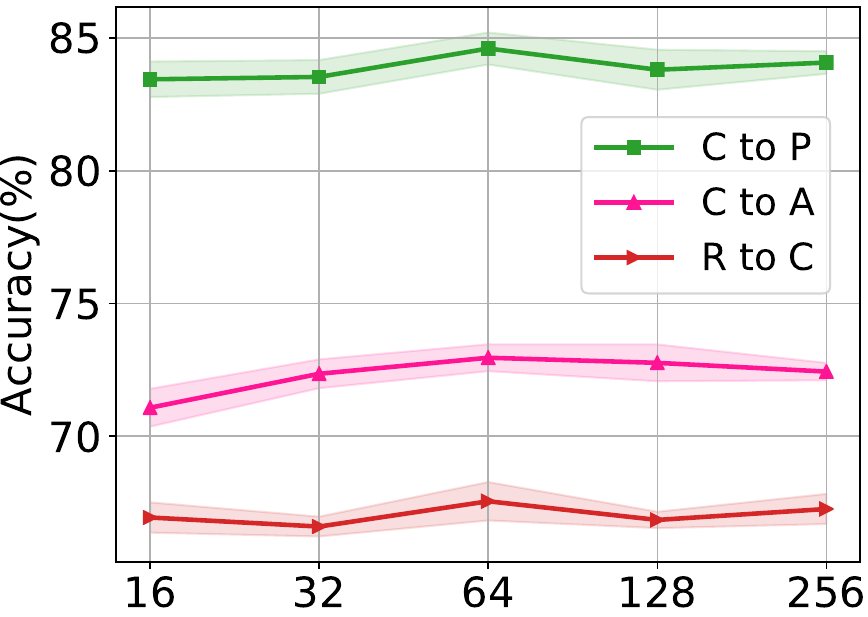}
    }
    \hspace{-3mm}
    \subfloat[Analysis of $\tau$]{
    \label{fig:sense-lambda}
    \includegraphics[width=0.233\textwidth]{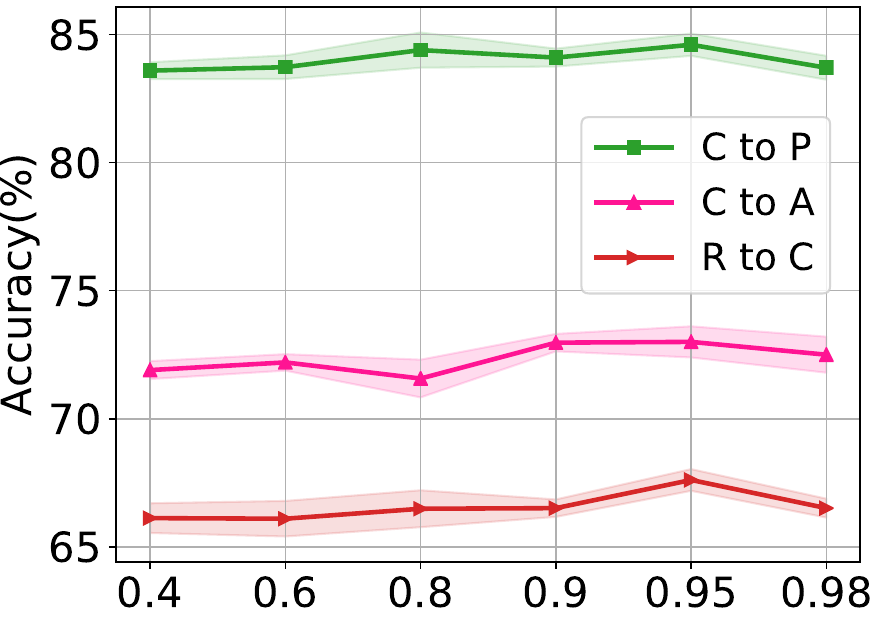}
    }
    \caption{Sensitivity with respect to the hyperparameters of \shortname{}.}
    \label{fig:sensi}
    \vspace{-5mm}
\end{figure}

We further carry out investigations to check the sensitivity of the proposed approach to the key hyper-parameters $\tau$ and $k$. We conduct the experiments in three diverse adaptation scenarios with varying degrees of transferring difficulty, namely C$\to$P, C$\to$A, and R$\to$C, on Office-Home. In Figure \ref{fig:sensi}, we show the test accuracy of the final adaptation model to display its classification performance when we set $\tau$ and $k$ to $\{0.40, 0.60, 0.80, 0.90, 0.95, 0.98\}$ and $\{16, 32, 64, 128, 256\}$, respectively. As illustrated, the trained model with $\tau=0.95$ and $k=64$ together can achieve a relatively higher classification performance than that of other cases, demonstrating the excellent choice of setting both hyper-parameters to 0.95 and 64, respectively.

\end{document}